\documentclass[a4paper,fleqn,svgnames,dvipsnames,table]{cas-dc}
\usepackage{stfloats}

\usepackage[numbers, sort&compress]{natbib}
\usepackage{caption}
\usepackage{bbding} 
\usepackage{amsmath}
\usepackage{hyperref}

\usepackage{xcolor}     
\newcommand{\first}[1]{\textcolor{red}{#1}}
\newcommand{\second}[1]{\textcolor{blue}{\underline{#1}}}
\newcommand{\third}[1]{\textcolor{orange}{\textit{#1}}}

\usepackage{graphicx}
\usepackage{bbding}
\usepackage{amsmath}
\usepackage{amsfonts}
\usepackage{enumerate}
\usepackage{booktabs}
\usepackage{algorithm}
\usepackage{algorithmic}
\usepackage{color}
\usepackage{colortbl}
\usepackage{bbm}
\usepackage{multirow}
\usepackage{array}
\usepackage{pifont}
\usepackage{xcolor}

\definecolor{greenc}{RGB}{0,176,80}

\makeatletter
\newcommand{\removelatexerror}{\let\@latex@error\@gobble}
\makeatother

\def\tsc#1{\csdef{#1}{\textsc{\lowercase{#1}}\xspace}}
\tsc{WGM}
\tsc{QE}

\begin{document}
\let\WriteBookmarks\relax
\def\floatpagepagefraction{1}
\def\textpagefraction{.001}

\shorttitle{}    
\shortauthors{Wang et al.}  

\title[mode = title]{
Multi-Modal Building Change Detection for Large-Scale Small Changes: Benchmark and Baseline}



\author[1]{Ye Wang$^\dagger$}[style=chinese]
\ead{wye@stu.ahu.edu.cn}

\author[1]{Wei Lu$^\dagger$}[style=chinese]   
\ead{luwei_ahu@qq.com}

\author[2]{Zhihui You}[style=chinese]   
\ead{youzh@aust.edu.cn}

\author[3]{Keyan Chen}[style=chinese]
\ead{keyan.chen@ntu.edu.sg}

\author[4]{Tongfei Liu}[style=chinese]   
\ead{liutongfei_home@hotmail.com}

\author[5]{Kaiyu Li}[style=chinese]   
\ead{likyoo.ai@gmail.com}

\author[6]{Hongruixuan Chen}[style=chinese]
\ead{qschrx@gmail.com}

\author[1]{Qingling Shu}[style=chinese]   
\ead{sql@stu.ahu.edu.cn}

\author[1]{Sibao Chen}[style=chinese]
\cormark[1]
\ead{sbchen@ahu.edu.cn}


\address[1]{MOE Key Lab of ICSP, Anhui Provincial Key Lab of Multimodal Cognitive Computation, IMIS Lab of Anhui Province, School of Computer Science and Technology, Anhui University, Hefei, China}
\address[2]{School of Public Safety and Emergency Management, Anhui University of Science and Technology, Hefei 231131, China}

\address[3]{College of Computing and Data Science, Nanyang Technological University, Singapore 639798}

\address[4]{Shaanxi Joint Laboratory of Artificial Intelligence, Shaanxi University of Science and Technology,
Xi’an 710021, China}
\address[5]{School of Software Engineering, Xi’an Jiaotong University, Xi’an 710049, China}

\address[6]{Graduate School of Frontier Sciences, The University of Tokyo, Chiba, 277-8561, Japan}

\cortext[1]{Corresponding author, $^\dagger$ Equal contribution.}

\begin{abstract}
Change detection in optical remote sensing imagery is susceptible to illumination fluctuations, seasonal changes, and variations in surface land-cover materials. Relying solely on RGB imagery often produces pseudo-changes and leads to semantic ambiguity in features. Incorporating near-infrared (NIR) information provides heterogeneous physical cues that are complementary to visible light, thereby enhancing the discriminability of building materials and tiny structures while improving detection accuracy. However, existing multi-modal datasets generally lack high-resolution and accurately registered bi-temporal imagery, and current methods often fail to fully exploit the inherent heterogeneity between these modalities. To address these issues, we introduce the Large-scale Small-change Multi-modal Dataset (LSMD), a bi-temporal RGB–NIR building change detection benchmark dataset targeting small changes in realistic scenarios, providing a rigorous testing platform for evaluating multi-modal change detection methods in complex environments. Based on LSMD, we further propose the Multi-modal Spectral Complementarity Network (MSCNet) to achieve effective cross-modal feature fusion. MSCNet comprises three key components: the Neighborhood Context Enhancement Module (NCEM) to strengthen local spatial details, the Cross-modal Alignment and Interaction Module (CAIM) to enable deep interaction between RGB and NIR features, and the Saliency-aware Multisource Refinement Module (SMRM) to progressively refine fused features. Extensive experiments demonstrate that MSCNet effectively leverages multi-modal information and consistently outperforms existing methods under multiple input configurations, validating its efficacy for fine-grained building change detection. The source code will be made publicly available at \url{https://github.com/AeroVILab-AHU/LSMD}. 
\end{abstract}
\maketitle
\begin{sloppypar}

\section{Introduction}\label{introduction}
Remote sensing change detection (RSCD) identifies surface variations by analyzing bi-temporal or multi-temporal images. It has been extensively applied in diverse fields, such as urban planning, land resource management, disaster assessment, and environmental monitoring \cite{A_shafique2022deep,A_wang2024advances}. Driven by the rapid acceleration of global urbanization, building-oriented change detection, encompassing new construction, demolition, and structural expansion, has emerged as a pivotal and challenging research task in the domain of high-resolution remote sensing interpretation \cite{A_cheng2024change}.

Recently, deep Convolutional Neural Networks (CNNs) and Transformer architectures have remarkably enhanced the representation of spatial features~\cite{LU2026431, A_SAAN,10142024}, thereby significantly improving the accuracy of automated building identification. Nevertheless, mainstream research predominantly focuses on single-modal change detection (SMCD), which typically relies solely on RGB imagery~\cite{A_LEVIR,A_WHU}. Given that RGB sensors are restricted to the visible spectrum, their performance is inherently susceptible to variations in illumination, atmospheric conditions, and seasonal vegetation phenology. In intricate urban scenes, the prevalence of intra-class spectral variability (e.g., identical building materials under varying lighting or viewpoints) and inter-class spectral similarity (e.g., color-proximate roofs and bare land) introduces substantial interpretation uncertainty. Consequently, single-modal models encounter severe information bottlenecks in suppressing pseudo-changes and extracting robust features under complex backgrounds.

\begin{figure*}[t]  
    \centering
   \includegraphics[width=1\textwidth]{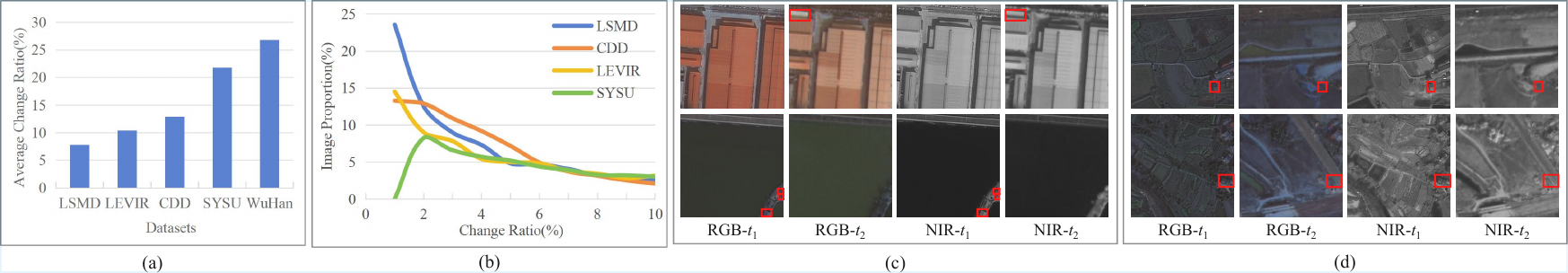} 
    \caption{Illustration of data distribution and challenging scenarios in realistic change detection. Note: unchanged images are excluded in (a) and (b) to focus on change-relevant statistics.
(a) Comparison of the average change ratio per image between the proposed LSMD and other mainstream benchmarks.
(b) Comparison of image proportions across different change ratios (1\%--10\%) between the proposed LSMD and existing benchmarks.
(c) Visual examples of small-scale changes in large-scale scenes.
(d) Visual examples of small buildings under vegetation backgrounds.}
    \label{NIR_dataset}
\end{figure*}

\begin{figure}
	\centering
	\includegraphics[width=0.95\linewidth]{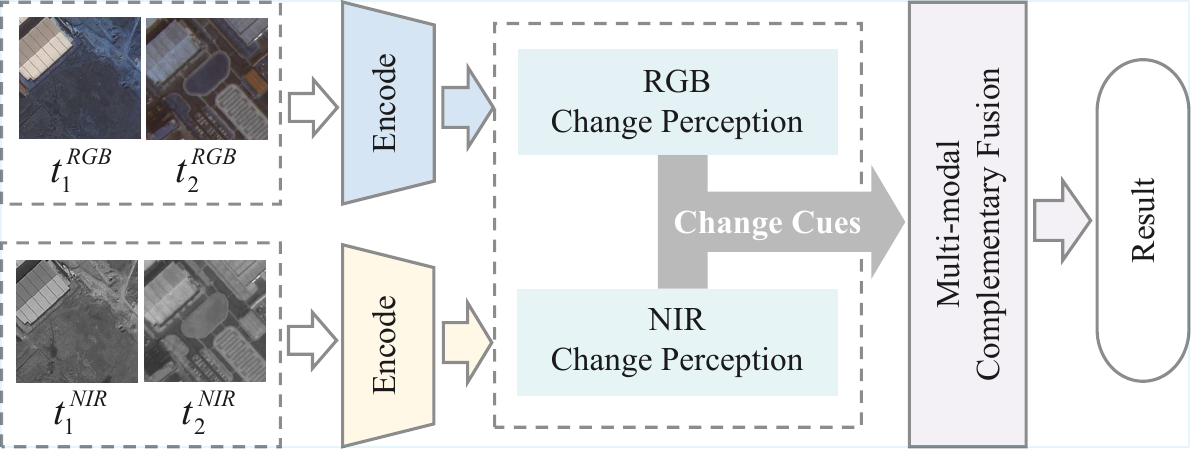}
	\caption{General architectural paradigm for MCD. Synchronous multi-modal imagery is provided at each temporal phase ($t_1$ and $t_2$). The architecture employs dual-branch feature encoding and independent change perception to capture initial temporal difference cues for each modality. Finally, a Multi-modal Complementary Fusion mechanism is utilized for the synergistic integration of multi-source information, generating highly robust change detection results.}
	\label{fig}
\end{figure}

To mitigate the spectral limitations of single-modal data, multi-modal information fusion has emerged as a prominent research focus in the field of RSCD. Based on the temporal modal configurations, relevant tasks are typically categorized into Heterogeneous Change Detection (HCD) and Multi-modal Change Detection (MMCD). Current benchmark datasets and methodologies are predominantly oriented toward HCD, which involves bi-temporal imagery captured by distinct sensors~\cite{A_xiongan,A_MT-wuhan}. Although HCD enhances change perception to some extent, the inherent disparities in physical imaging mechanisms and temporal asynchrony pose significant challenges for precise cross-modal feature alignment and correlation modeling. In contrast, MMCD provides multi-modal observations at each time point~\cite{A_3DCD,A_SMARS}, establishing a more direct data foundation for joint feature modeling. However, research in the MMCD domain remains relatively underdeveloped, characterized by a scarcity of public datasets and limitations in spatial resolution or modal diversity, which constrains the capacity for fine-grained modeling in intricate urban environments.

Existing multi-modal efforts frequently augment RGB imagery with Synthetic Aperture Radar (SAR) or Digital Surface Models (DSM) to incorporate complementary geometric or topographic information~\cite{A_SAR, A_DSM}. However, in practical applications, the inherent speckle noise and geometric distortions of SAR imagery exacerbate the difficulty of cross-modal feature alignment~\cite{A_SAR_juxian}. Meanwhile, the acquisition of high-quality DSMs is characterized by high costs and protracted update cycles, which constrains their large-scale engineering deployment for dynamic monitoring~\cite{A_DSM_juxian}. By contrast, Near-Infrared (NIR) bands achieve inherent geometric congruence with RGB through synchronous acquisition, effectively reducing cross-modal alignment complexity and minimizing fusion errors induced by geometric distortions. Furthermore, the significant differences between the two modalities in physical response mechanisms and reflectance characteristics provide vital spectral complementarity, laying a rich foundation for constructing synergetic multi-modal representations and serving as a more reliable source for robust urban change detection.

Furthermore, in the construction of existing change detection benchmark datasets, image patches with a high proportion of changed areas are often deliberately selected to ensure sample efficacy. Such manual filtering leads to a data distribution characterized by "high density and prominent features," which deviates from the objective reality of wide-area remote sensing monitoring, where changed regions are typically characterized by extreme spatial sparsity. Consequently, while models may demonstrate superior performance in controlled laboratory settings, their generalization capability and robustness are significantly constrained when deployed in real-world complex scenarios.

To better align with practical remote sensing monitoring conditions, this study prioritizes two realistic challenges during dataset construction and methodology design. First, the detection of subtle changes within large-scale scenes. In real-world wide-area imagery, change patterns typically exhibit a distribution characteristic of "expansive backgrounds with sparse variations." Compared to the relatively high average change proportions found in mainstream benchmark datasets (see Fig.~\ref{NIR_dataset}a), such high-density distributions may lead models to over-rely on strong spatial priors to learn change patterns. In contrast, during our dataset construction, the proportion of changed regions is strictly controlled, with an increased emphasis on small-scale change samples, particularly those with a change ratio below 2\% (as illustrated in Fig.~\ref{NIR_dataset}b). This setup simulates the extreme class imbalance and spatial sparsity encountered in real-world scenarios, significantly raising the difficulty threshold for the detection task. It necessitates models capable of precisely locating minute targets within vast and complex backgrounds, as visualized in Fig.~\ref{NIR_dataset}c. Second, subtle building changes under vegetation cover. During practical urbanization, numerous new small-scale buildings are constructed directly on natural surfaces such as grasslands, woodlands, or farmlands (see Fig.~\ref{NIR_dataset}d). These targets often lack distinct contextual structural information in RGB imagery and are easily obscured by complex backgrounds. From the perspective of spectral physical response, healthy vegetation exhibits high reflectance in the NIR band, whereas typical construction materials (e.g., concrete and metal) manifest significantly different radiative signatures. Consequently, when natural vegetation is replaced by buildings, the NIR response undergoes a sharp physical transition. Motivated by these considerations, we constructed the Large-scale Small-change Multi-modal Dataset (LSMD), comprising accurately registered bi-temporal RGB-NIR imagery, to address these two critical challenges in realistic remote sensing scenarios.

To fully exploit the multi-modal spectral information within the LSMD dataset and address the research void in the MMCD field, we propose a novel architectural paradigm for MCD, as illustrated in Fig.~\ref{fig}. This paradigm is designed to efficiently process synchronized multi-modal inputs at each time step, deeply integrating the complementary advantages of diverse modalities through independent feature encoding and cross-modal cue extraction. Under this paradigm, we further develop the Multi-modal Spectral Complementarity Network (MSCNet), which leverages multi-level interaction mechanisms to harness the synergistic gains from spatial structures and spectral responses. This integration specifically enhances detection performance for concealed small buildings and subtle change regions. Specifically, a Neighborhood Context Enhancement Module (NCEM) is introduced to strengthen local contextual perception via multi-scale neighborhood interactions. Considering the distinct statistical properties of RGB and NIR data, a Cross-modal Alignment and Interaction Module (CAIM) is designed to perform adaptive feature calibration and deep cross-modal interaction on bi-temporal differential features. Furthermore, to suppress pseudo-changes and bolster the representation of changed regions, we propose a Saliency-aware Multi-source Refinement Module (SMRM). Rather than directly integrating large foundation models into the inference pipeline, the SMRM adopts an offline strategy by utilizing semantic masks generated from a pre-trained segmentation model (RemoteSAM~\cite{A_RemoteSAM}) as spatial priors, thereby avoiding additional computational overhead during inference. Guided by these masks and multi-modal differential features, the SMRM progressively refines the fusion representations to enhance change saliency and ensure robustness in intricate urban scenes. The primary contributions of this work are summarized as follows:
\begin{itemize}
\item[1)] We introduce LSMD, a bi-temporal MMCD dataset comprising 8,000 pairs of high-resolution images. Each temporal phase provides accurately registered RGB and NIR data, with a particular focus on small changes in large-scale real-world remote sensing scenes, offering a unified foundation for advancing MMCD research in complex urban scenarios.
\item[2)] We propose MSCNet, which implements synergistic multi-modal feature learning through three core modules. Specifically, the NCEM strengthens local spatial details, the CAIM facilitates the deep integration of RGB and NIR features, and the SMRM refines fusion representations, collectively enhancing detection performance for complex urban targets.
\item[3)] Experimental results on the LSMD and SMARS datasets illustrate that the proposed method achieves superior performance over existing change detection methods, further confirming the value of multi-modal collaborative feature learning.
\end{itemize}

\section{Related Work} \label{Section2}
\subsection{RSCD Datasets}
Most change detection studies are based on SMCD datasets, which provide high-resolution optical imagery at each temporal phase, such as LEVIR~\cite{A_LEVIR}, WHU~\cite{A_WHU}, CDD~\cite{A_CDD}, and SYSU~\cite{A_SYSU}. However, reliance on single-spectral information makes these datasets inherently vulnerable to illumination-induced pseudo-changes and phenological variations in complex environments.

To overcome the limitations of single-modal optical data, researchers have introduced HCD and MMCD datasets. Representative HCD datasets include XiongAn~\cite{A_xiongan},  MT-Wuhan~\cite{A_MT-wuhan} and BRIGHT~\cite{Chen2025BRIGHTAG}, which incorporate cross-modal information to enhance change perception to some extent~\cite{A_RCAM,LIU2025111355}; however, due to the imaging mechanism discrepancies and inherent geometric misalignment between disparate sensors, achieving precise spatial-level registration remains a formidable challenge. In contrast, MMCD datasets, such as 3DCD~\cite{A_3DCD}, SMARS~\cite{A_SMARS}, and MOSCD~\cite{A_OSCD}, provide synchronized multi-modal observations at each temporal phase, supporting joint feature modeling and accurate difference analysis; nevertheless, these datasets remain limited in spatial resolution, sample scale, or modality combinations, making fine-grained building change modeling in complex urban environments difficult.

In contrast, this work constructs the LSMD dataset, a high-resolution, bi-temporal multi-modal change detection benchmark based on an RGB–NIR multispectral configuration, covering real urban scenes and focusing on fine-grained building changes in large-scale backgrounds, with unified imaging geometry. Since the data are synchronously acquired by the same sensor, pixel-level precise registration is ensured, while highly complementary spectral information is provided, making the dataset particularly suitable for high-resolution building change detection tasks.

\subsection{SMCD and HCD Methods} 
Most SMCD methods rely on differencing-based networks or Siamese architectures to compare bi-temporal features, often incorporating context modeling, attention mechanisms, and multi-scale strategies to enhance performance. For example, BiFA~\cite{A_BiFA} mitigates illumination and perspective differences via bi-temporal feature alignment, while T-UNet~\cite{A_UNet} aggregates bi-temporal and fused features within a UNet-based encoder–decoder to alleviate feature distribution discrepancies. Although these methods achieve strong performance on RGB benchmarks, their applicability in heterogeneous or multi-modal scenarios remains limited.

HCD methods primarily focus on modeling feature inconsistencies across different modalities. For instance, GLCD-DA~\cite{A_GLCD} maps optical and SAR images into a unified feature space using an image translation network, incorporating global–local interaction and multi-level feature fusion to enhance change representation. CFRL~\cite{10891329} further alleviates cross-modal discrepancies through feature representation learning and hierarchical feature alignment. However, these methods generally assume a single modality at each temporal phase and mainly rely on modality mapping or feature alignment, limiting their ability to fully exploit complementary multi-modal information.

\subsection{Multi-modal Interaction and Semantic Prior}
Due to the limited availability of MMCD datasets, research in this area remains relatively scarce, and existing methods mainly focus on strongly heterogeneous modality combinations, such as optical–SAR scenarios. For instance, MSCD-Net~\cite{A_MSCD-Net} performs change detection by fusing optical and SAR features while jointly modeling semantic and difference information, and HF-MCD~\cite{A_HF-MCD} leverages HCF and HAF modules to mitigate resolution differences and feature inconsistencies. Nevertheless, fine-grained change detection between homologous optical modalities remains challenging, particularly when it comes to capturing subtle variations in local building structures and material properties.

Single-temporal multi-modal fusion methods demonstrate strong feature modeling capabilities. For example, HGN~\cite{A_HGN} achieves progressive fusion at both feature and decision levels through hierarchical gating mechanisms; MCTUNet~\cite{A_MCTUNet} combines CNN and Transformer architectures to enhance local–global interactions between optical and elevation modalities; and MDFNet~\cite{A_MDFNet} decomposes features into shared and modality-specific components to alleviate modality inconsistencies. However, these methods focus on static intra-temporal aggregation, neglecting the inter-temporal modeling needed to distinguish genuine transitions from temporal noise.

For bi-temporal RGB–NIR change detection, models are required not only to effectively integrate cross-modal complementary information, but also to explicitly model temporal evolution to distinguish genuine land-cover changes from pseudo-changes caused by radiometric discrepancies or phenological fluctuations. Existing public datasets and methods, however, still lack systematic investigation and evaluation frameworks tailored to this task, which restricts method development and fair comparison. To address these issues, this work proposes a unified framework for bi-temporal RGB–NIR multi-modal change detection and further introduces offline semantic priors inspired by large-scale foundation models, which enable more refined and accurate modeling of land-cover changes.

Recent studies have begun incorporating large-scale foundation models into remote sensing interpretation tasks to leverage their strong semantic understanding capability~\cite{11313649, Li2025AnnotationFreeOS, li2024semicdvlvisuallanguagemodelguidance}. Among these models, the Segment Anything Model (SAM) has received considerable attention for its remarkable zero-shot segmentation ability, providing effective support for constructing semantic priors in change detection. Methods such as TTP~\cite{A_TTP}, RCAM~\cite{A_RCAM}, and FAEWNet~\cite{A_FAEWNet} have successfully leveraged SAM-generated semantic priors, improving the accuracy of target localization. However, directly embedding SAM into the online inference pipeline often incurs substantial computational overhead. To address this issue, the proposed approach adopts an offline strategy by introducing RemoteSAM~\cite{A_RemoteSAM}, which is specifically designed for remote sensing scenarios and contains only 180M parameters. By precomputing its generated masks as spatial priors, the proposed method enables precise target focusing and effective suppression of pseudo-changes, while avoiding additional computational burden during online inference.

\begin{figure*} 
    \centering
   \includegraphics[width=1\textwidth]{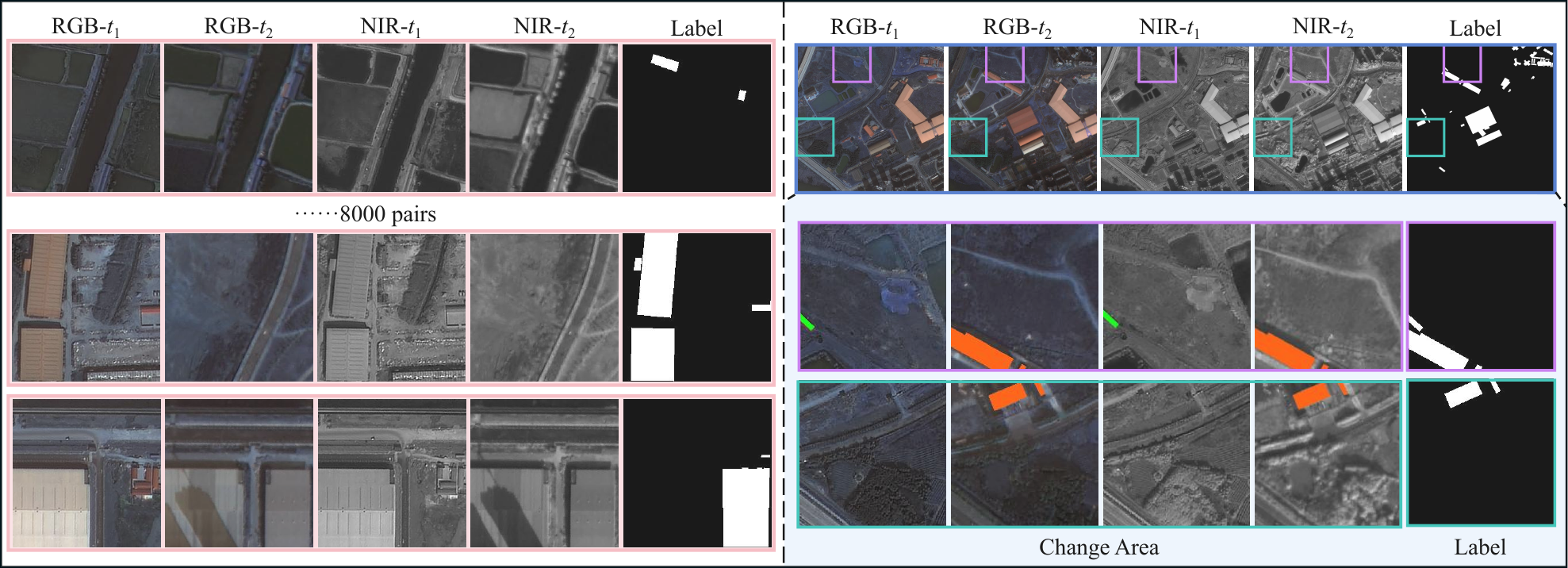} 
    \caption{Samples and annotations from the LSMD dataset. (Left) Bi-temporal RGB and NIR image pairs with ground truth. (Right)  Detailed illustrations of changed regions.}
    \label{fig_1}
\end{figure*}

\begin{table*}[t]
\centering
\small
\caption{Summary of widely used change detection datasets}
\label{table_1}
\renewcommand{\arraystretch}{1.25} 
\setlength{\tabcolsep}{3.5pt}         
 \begin{tabular}{l c c c c c c}
\hline
\textbf{Dataset} & 
\textbf{Resolution} &
\textbf{Image pairs} &
\textbf{Image size (pixels)} &
\textbf{$t_1$ type} &
\textbf{$t_2$ type} &
\textbf{Objects of interest} \\
\hline

\multicolumn{7}{l}{\textbf{SMCD Dataset}} \\
\hline
LEVIR\cite{A_LEVIR}  & 0.5m   & 637   & $1,024 \times 1,024$   & RGB & RGB & Building \\
WHU\cite{A_WHU}      & 0.2m   & 1     & $32,207 \times 15,354$ & RGB & RGB & Building \\
CDD\cite{A_CDD}        & 0.03m--1m & 16,000 & $256 \times 256$     & RGB & RGB & Multi-class \\
SYSU\cite{A_SYSU}       & 0.5m   & 20,000 & $256 \times 256$     & RGB & RGB & Building \\

\hline
\multicolumn{7}{l}{\textbf{HCD Dataset}} \\
\hline
CAU-Flood\cite{HE2023103197} & 10m    & 18,302     & $256 \times 256$  & Optical & SAR & Flooding \\
XiongAn\cite{A_xiongan}    & 4m/8m & 2,360  & $512 \times 512$   & RGB & SAR & Building \\
MT-Wuhan\cite{A_MT-wuhan}   & 10m/3m & 1     & $11,216 \times 13,693$ & RGB & SAR & Building \\
BRIGHT\cite{Chen2025BRIGHTAG}   & 0.3m--1m & 4,246     & $1,024 \times 1,024$ & RGB & SAR & Building damage  \\

\hline
\multicolumn{7}{l}{\textbf{MMCD Dataset}} \\
\hline
3DCD\cite{A_3DCD}     & 0.5m/1.0m & 472 & RGB: $400 \times 400$/DSM: $200 \times 200$ & RGB+DSM & RGB+DSM & Building \\
SMARS\cite{A_SMARS}    & 0.3m       & 2 (Scenes) & \makecell{SVenice: $5,600 \times 5,600$ \\ SParis: $5,600 \times 5,600$} & RGB+DSM & RGB+DSM & Building \\
SMARS\cite{A_SMARS}    & 0.5m       & 2 (Scenes) & \makecell{SVenice: $5,600 \times 5,600$ \\ SParis: $4,500 \times 3,560$} & RGB+DSM & RGB+DSM & Building \\
MOSCD\cite{A_OSCD} & 10m--20m   & 24 & Variable & Optical+SAR & Optical+SAR & Urban \\

\hline
LSMD(Ours)       & 0.8m & 8,000 & $256 \times 256$ & RGB+NIR & RGB+NIR & Building \\
\hline
\end{tabular}
\end{table*}

\section{LSMD Dataset}\label{Section3}

The study area encompasses selected regions in Wuhan, China, representing a typical urban-rural composite landscape characterized by a diverse array of complex land cover types, including buildings, roads, villages, farmlands, and vegetation. Bi-temporal high-resolution images acquired by the Gaofen-2 (GF-2) satellite in 2016 and 2023 were utilized as the primary data source. Driven by the significant urbanization process within this interval, extensive real-world structural changes in buildings have occurred, providing a wealth of reliable samples for the construction of a large-scale building change detection dataset.

Change annotations were generated through manual delineation of regions of interest. To meet the input requirements of deep learning models, the original images were subjected to radiometric correction, geometric registration, and normalization. Subsequently, the images were cropped using fixed-size sliding windows to yield paired RGB–NIR samples. To address the challenges of large-scale scenes and small changes in wide-area remote sensing monitoring, we deliberately increased the proportion of subtle building change samples when constructing the dataset to ensure diversity and representativeness. Ultimately, the LSMD dataset comprises a total of 8,000 bi-temporal multi-modal samples with a spatial size of 
256 $\times$ 256 pixels, where each temporal phase of a sample contains independent RGB and NIR modalities. These samples are partitioned into a training set (5,598 pairs), a validation set (796 pairs), and a test set (1,606 pairs) according to a ratio of 7:1:2. Representative annotation examples are illustrated in Fig.~\ref{fig_1} to provide a clear and intuitive visual reference.

Table~\ref{table_1} provides a systematic comparison between the proposed dataset and existing mainstream change detection datasets. Existing SMCD datasets rely solely on RGB imagery, which offers limited spectral information. HCD datasets typically involve the fusion of optical imagery and SAR data, where significant modality discrepancies make feature alignment and fusion extremely difficult. Among existing MMCD datasets, MOSCD suffers from low spatial resolution (10–20 m), limiting its applicability for fine-grained building change detection. While the SMARS dataset provides high-resolution images (0.3 m and 0.5 m), it is essentially synthetic, making it challenging to fully capture the complex texture characteristics and noise properties of real-world remote sensing scenes. Although the 3DCD dataset explores multi-modal fusion, its sample size is relatively small, and cross-modal resolution inconsistencies further increase the difficulty of feature alignment.

Consequently, there is still a lack of a multi-modal benchmark dataset that combines large-scale samples, high spatial resolution, and real-world physical characteristics, while providing both pixel-level geometric consistency and rich feature information. LSMD is specifically designed to fill this gap; by leveraging the significant heterogeneity in information representation between RGB and NIR, it provides more reliable information support while maintaining high spatial resolution. Its inherent pixel-level geometric consistency eliminates the registration errors common in heterogeneous data (such as SAR or DSM), ensuring precise localization of small-scale change targets. Furthermore, by utilizing the unique physical reflectance differences of NIR for vegetation and building materials, LSMD overcomes the issues of signal attenuation and detail loss in RGB caused by environmental factors. At the same time, it exploits the high sensitivity to chlorophyll to distinguish phenological changes from real land-cover evolution, significantly enhancing the overall robustness and reliability of detection under non-ideal imaging conditions.

\section{Methodology}\label{Section4}
\subsection{Overview}
\begin{figure*}[!t]  
    \centering
    \includegraphics[width=1\textwidth]{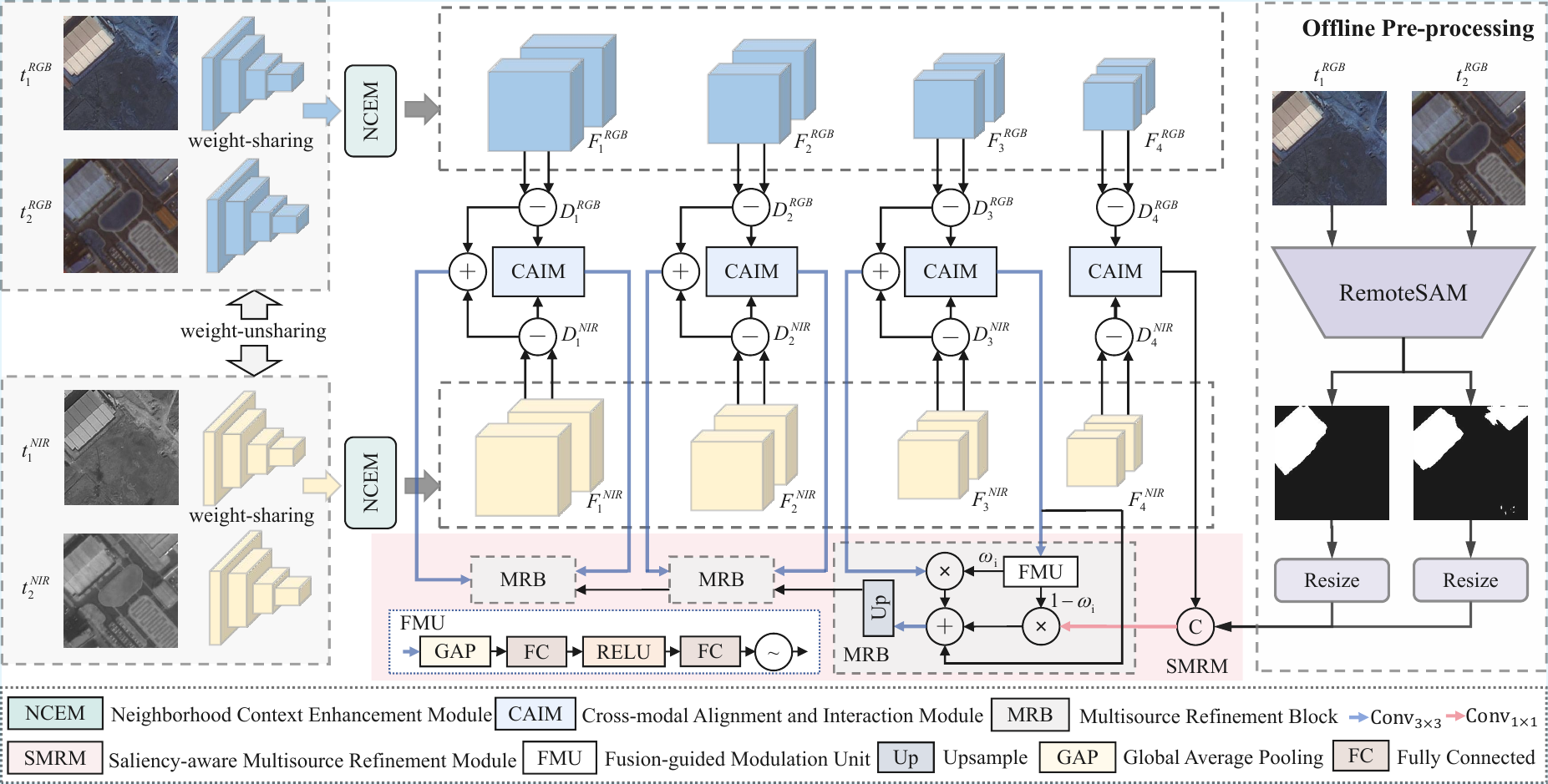} 
    \caption{Overall architecture of the proposed MSCNet. First, a Siamese backbone is employed to extract bi-temporal RGB and NIR features separately. Next, the Neighborhood Context Enhancement Module (NCEM) enhances the features to capture local change information. The Cross-modal Alignment and Interaction Module (CAIM) then integrates the RGB and NIR difference features to generate a more discriminative fused representation. Subsequently, the Saliency-aware Multisource Refinement Module (SMRM) leverages high-level semantic priors generated offline by RemoteSAM and, under the guidance of semantic masks, multi-modal difference features, and multi-scale context, progressively refines the fused features to restore spatial resolution and maintain cross-scale feature consistency. Finally, the features processed by these modules are used to generate the change detection results.}
    \label{fig_2}
\end{figure*}
As shown in Fig.~\ref{fig_2}, the proposed MSCNet takes bi-temporal RGB and NIR images as input, with two temporal phases ($t_1$ and $t_2$) for each modality. Feature extraction is conducted by a four-branch MobileNetV2 backbone. A weight-sharing strategy is adopted for images of the same modality (e.g., $t_1^{RGB}$ and $t_2^{RGB}$) to ensure feature consistency, while weight unsharing is applied across modalities to accommodate distinct spectral distributions. Multi-level feature maps from the last four stages of MobileNetV2 are extracted and fed into Neighborhood Context Enhancement Module (NCEM), which enhances local detail representation while preserving contextual consistency for each modality at each temporal phase, where the feature enhancement process is formulated as follows:
\begin{equation}
\begin{split}
    F_{i, t_n}^{m} = \mathrm{NCEM}_i(f_{i, t_n}^{m}), \quad i &\in \{1,2,3,4\}, n \in \{1,2\}, \\
    m &\in \{RGB, NIR\},
\end{split}
\end{equation}
where $f_{i, t_n}^{m}$ denotes the raw feature map output from the backbone network at the $i$-th level for modality $m$ at temporal phase $t_n$, and $F_{i, t_n}^{m}$ represents the enhanced feature at the $i$-th level for modality $m$ at temporal phase $t_n$. 

To emphasize inter-temporal changes, element-wise subtraction is applied between enhanced features from different temporal phases within each modality. The modality-specific difference representations are computed as:
\begin{equation}
\begin{split}
    D_i^m = \left| F_{i,t_1}^m \ominus F_{i,t_2}^m \right|, \quad i &\in \{1, 2, 3, 4\}, \\
    m &\in \{RGB, NIR\},
\end{split}
\end{equation}
where $|\cdot|$ denotes the absolute value operation, $\ominus$ denotes the element-wise subtraction operation, $F_{i,t_1}^m$ and $F_{i,t_2}^m$ are the enhanced features at $t_1$ and $t_2$, respectively, and $D_i^m$ represents the difference feature at level $i$ for modality $m$.

The modality-specific difference features are then fed into Cross-modal Alignment and Interaction Module (CAIM) for cross-modal interaction and complementary enhancement. The multi-modal feature fusion at each level is defined by:
\begin{equation}
    M_i = \mathrm{CAIM}\left( D_i^{RGB}, D_i^{NIR} \right), \quad i \in \{1,2,3,4\},
\end{equation}
where $M_i$ denotes the fused multi-modal feature at the $i$-th level. Next, saliency masks $\text{mask}_{t_1}$ and $\text{mask}_{t_2}$, obtained offline from a pre-trained segmentation model (RemoteSAM~\cite{A_RemoteSAM}), are incorporated as spatial priors. These masks provide strong semantic object distribution information, guiding the model to focus on actual building regions and facilitating saliency-aware contextual refinement. The acquisition process of these masks is formulated as follows:
\begin{align}
\mathrm{mask}_{t_i} &= \mathrm{RemoteSAM}(t_i^{RGB}), 
\quad i \in \{1,2\}, 
\end{align}
where $\text{mask}_{t_i}$ is the mask of the original RGB image at time $t_i$ obtained using RemoteSAM. Subsequently, the proposed Saliency-aware Multisource Refinement Module (SMRM) is employed to perform multi-scale saliency-guided refinement. Guided by the semantic masks and multi-modal difference features, a hierarchical top-down aggregation strategy is utilized to progressively recover the spatial resolution, which can be expressed as follows:
\begin{equation}
    M_i' = \mathrm{SMRM}\!\left( \!M_i,\! M_{i+1}',\! mask_{t_1},\! mask_{t_2} \!\right)\!, \!\!\!\!\quad i \! \in \! \{\!1,\!2,\!3,\!4\},
\end{equation}
ultimately, the enhanced features $M_{1}'$ are upsampled to the input dimensions via $1\times1$ convolutions and bilinear interpolation, followed by a Sigmoid activation to yield the final change prediction map $Y$, which is mathematically expressed as follows:
\begin{equation}
    Y = \sigma\left( \mathrm{UP}\left( \mathrm{Conv}_{1\times1}(M_1') \right)  \right),
\end{equation}
where $\mathrm{UP}(\cdot)$ denotes bilinear upsampling for spatial resolution recovery and $\sigma$ denotes the Sigmoid function.

\subsection{NCEM}
\begin{figure}  
    \centering
    \includegraphics[width=\linewidth]{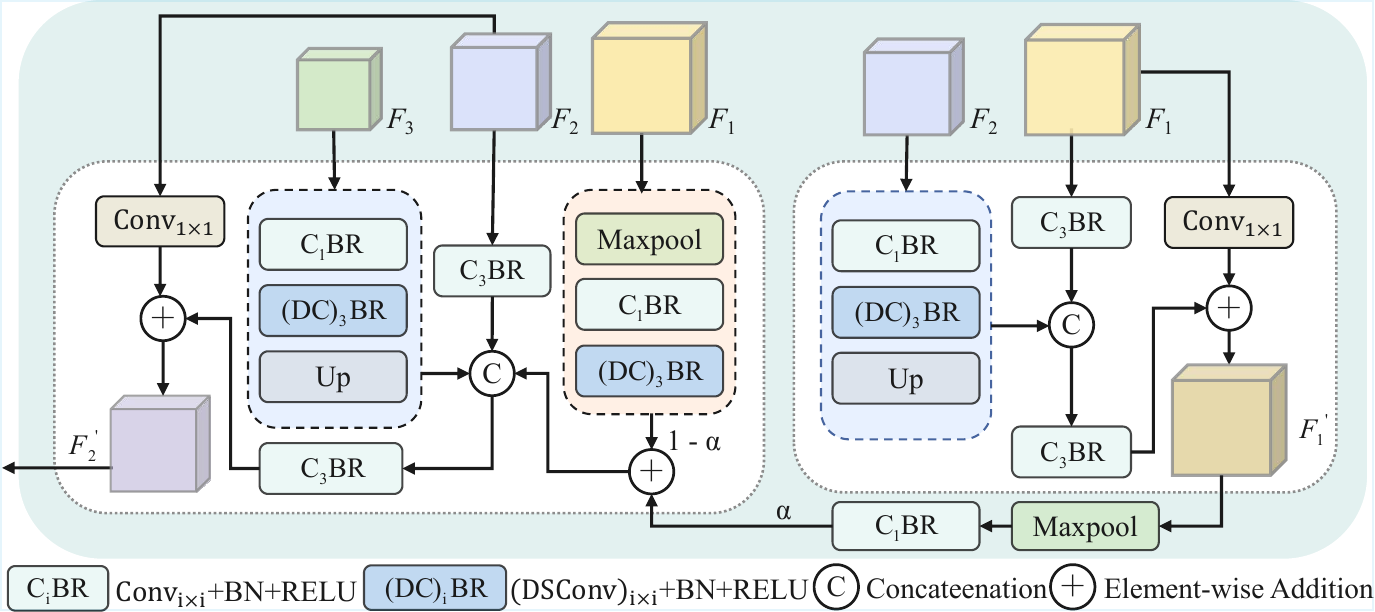} 
    \caption{Structure of NCEM. Multi-level neighboring features are selectively aggregated and adaptively weighted to enhance local spatial details and contextual consistency.}
    \label{fig_3}
\end{figure}

Motivated by the observation that low-level features lack sufficient semantic context while high-level features may miss fine spatial details, we introduce the Neighborhood Context Enhancement Module (NCEM), as illustrated in Fig.~\ref{fig_3}. NCEM enhances spatial details and preserves contextual consistency by selectively aggregating adjacent-layer features via a bottom-up neighbor propagation mechanism. To avoid redundancy from excessive fusion, it restricts interactions to neighboring layers and employs residual and selective neighborhood fusion strategies, thus maintaining local detail and contextual consistency.

Taking the second-stage feature as an example, $F_2$ serves as the primary feature, while the low-level feature $F_1$, the high-level feature $F_3$, and residual information from adjacent scales ($F_1'$) act as auxiliary features. Unless otherwise specified, batch normalization (BN) and ReLU activation are applied after all convolutional layers in NCEM. The primary feature $F_2$ is first processed by a $3\times3$ convolution to enhance semantic representation. The low-level feature $F_1$ is downsampled via max pooling, compressed by a $1\times1$ convolution, and refined with a $3\times3$ depthwise separable convolution (DW-Conv). Its residual feature $F_1'$ is downsampled and passed through a $1\times1$ convolution to align its channels with $F_1$, and then fused with the feature generated from $F_1$ using a learnable coefficient $\alpha$, achieving bottom-up neighbor enhancement. The high-level feature $F_3$ is processed through a $1\times1$ convolution and a $3\times3$ DW-Conv, and its resolution is upsampled to match that of $F_2$. The feature transformation is formulated as follows:
\begin{align}
\begin{split}
F_1^1 = & (1 - \alpha) \times (\mathrm{DSConv}_{3 \times 3} (\mathrm{Conv}_{1 \times 1} (\mathrm{MP} (F_1)))) \\
& + \alpha \times (\mathrm{Conv}_{1 \times 1} (\mathrm{MP} (F'_1))),
\end{split} \\  
F_2^1 = & \mathrm{Conv}_{3 \times 3}(F_2), \\[1ex]
F_3^1 = & \mathrm{UP}(\mathrm{DSConv}_{3 \times 3}(\mathrm{Conv}_{1 \times 1}(F_3))),
\end{align}
where $\mathrm{MP}(\cdot)$ represents the max pooling operation, $\alpha$ is a learnable parameter balancing the current features and residual features, and $\mathrm{DSConv}_{3 \times 3}$ represents a $3\times3$ DW-Conv; BN and ReLU activation are applied to most convolutional layers but are omitted in the formulas for simplicity.

After transformation, features are concatenated to aggregate multi-scale information. The primary feature $F_2$ is compressed via a $1\times1$ convolution without batch normalization and activation, which serves as a linear projection for residual fusion, while the fused feature undergoes a $3\times3$ convolution to enhance contextual difference representation. The final enhanced feature is obtained via a residual connection, producing contextually enhanced features that preserve the original representation and facilitate stable gradient propagation. This process is formulated as follows:
\begin{equation}
    F_2' = \text{Conv}_{3 \times 3}\left( \mathrm{Cat}\left[ F_1^1, F_2^1, F_3^1 \right] \right) + \mathrm{Conv}_{1\times1}(F_2).
\end{equation}

NCEM is also applicable to features at other stages. By adopting the same neighborhood enhancement and multi-scale feature aggregation strategy, enhanced features at each stage can be obtained for both RGB and NIR modalities, denoted as $F_1', F_2', F_3', F_4'$.
\subsection{CAIM}
\begin{figure*}[!t]  
    \centering
    \includegraphics[width=0.95\textwidth]{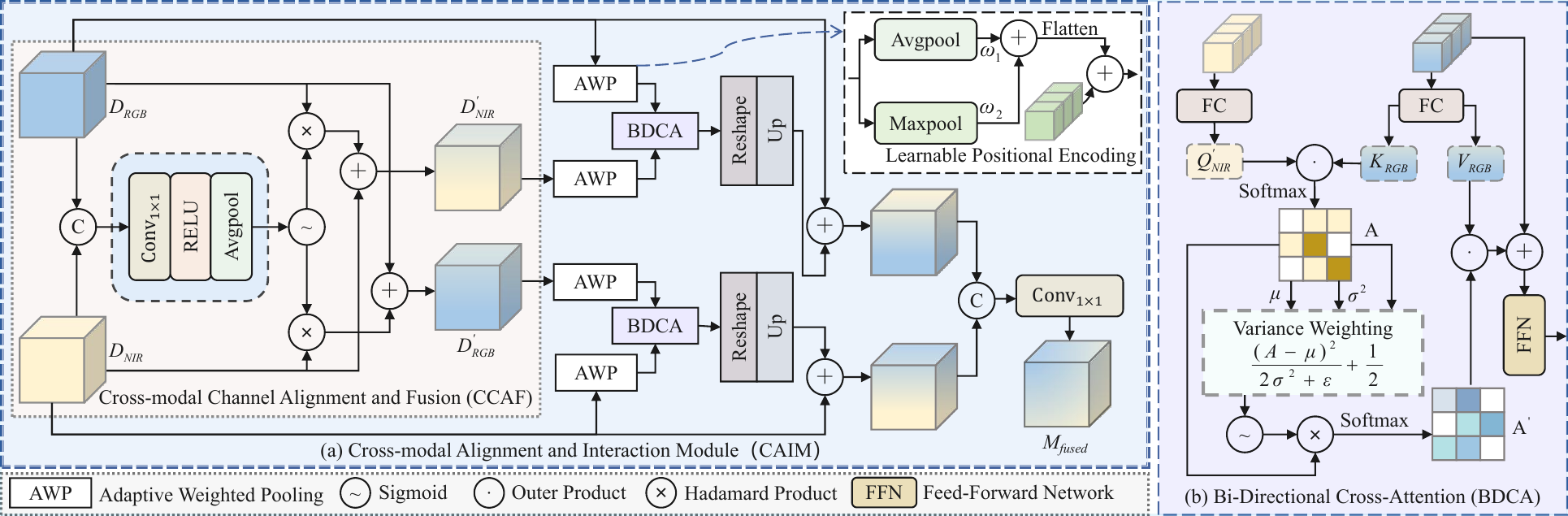} 
    \caption{Structure of CAIM. (a) Overall multi-modal fusion framework. (b) Bi-Directional Cross-Attention (BDCA). Cross-modal Channel Alignment and Fusion (CCAF), Adaptive Weighted Pooling (AWP), and BDCA are jointly employed to enable deep interaction between RGB texture features and NIR spectral features.}
    \label{fig_4}
\end{figure*}
Motivated by the complementary characteristics of RGB and NIR modalities, we propose the Cross-modal Alignment and Interaction Module (CAIM), as shown in Fig.~\ref{fig_4}. This module is designed to achieve cross-modal feature alignment and joint representation learning.

To alleviate the distribution discrepancy between RGB and NIR features, a Cross-modal Channel Alignment and Fusion (CCAF) mechanism is introduced. The difference features are first concatenated and processed by stacked $1\times1$ convolutions for joint channel compression and preliminary fusion. A channel attention module based on average pooling and Sigmoid activation is then applied to model inter-channel dependencies, producing attention weights:
\begin{equation}
    A_1 = \sigma\left( \mathrm{AP}\left( \delta\left( \mathrm{Conv}_{1\times1}\left( \mathrm{Cat}\left[ \ D_{RGB}, D_{NIR} \right] \right) \right) \right) \right),
\end{equation}
where $\delta$ denotes the ReLU activation function, and $\mathrm{AP}(\cdot)$ denotes the average pooling operation.

The generated weights are used to perform residual weighted fusion of the original modal features, producing aligned and fusion-enhanced representations. The enhanced features can be directly fed into subsequent cross-attention or convolutional operations to strengthen cross-modal representation learning. This enhancement is formulated as:
\begin{align}
    D_{RGB}' &= D_{RGB} + D_{NIR} \times A_1, \\
    D_{NIR}' &= D_{NIR} + D_{RGB} \times A_1,
\end{align}
to reduce computational complexity while incorporating multi-scale spatial context, an adaptive weighted pooling (AWP) strategy is applied to the aligned features, in which average pooling and max pooling are combined via learnable weights, and the resulting pooled features are flattened into token sequences augmented with learnable positional encodings to preserve spatial structure.

Subsequently, the flattened features are fed into the Bi-Directional Cross-Attention (BDCA) mechanism to enable deep interaction and mutual enhancement between RGB and NIR features. Taking the RGB feature as an example, it is first mapped to Keys ($K$) and Values ($V$), while the enhanced NIR feature is mapped to Queries ($Q$):
\begin{equation}
    V_{RGB}\!=\!T_{\!RGB}W_{\!V},\! K_{RGB}\!=\!T_{\!RGB}W_{\!K},\! Q_{NIR}'\!=\!T_{\!NIR}'W_{\!Q},
\end{equation}
where $W$ denotes the projection matrices. The cross-modal correlation is calculated through the dot product of $Q$ and $K$, and a preliminary cross-modal attention weight map $A$ is obtained via Softmax. This process can be formulated as:
\begin{equation}
 A_{RGB} = \mathrm{Softmax}\left( \frac{Q_{NIR}' K_{RGB}^T}{\sqrt{D_k}} \right),
\end{equation}
where $\sqrt{D_k}$ is the scaling factor.

To further suppress background noise and highlight building change regions, we introduce a variance weighting mechanism for attention refinement. In remote sensing imagery, regions where buildings undergo changes are typically accompanied by significant variations in spatial structure and spectral characteristics, which manifest as high-variance responses in the attention maps, whereas homogeneous background regions produce low-variance responses. Based on this observation, a variance-based modulation term is constructed to adaptively adjust the initial attention. Specifically, the variance weight term is generated based on the mean $\mu$ and variance $\sigma^2$ of the attention map $A$, thereby enhancing the model’s response to salient change regions. This weight is multiplied with the original attention map and then processed through the softmax function to obtain the enhanced attention weights $A'$.
\begin{equation}
    A'_{RGB} = \mathrm{Softmax}\!\left(\! A_{RGB} \times \sigma\left(\! \frac{(A_{RGB} - \mu)^2}{2\sigma^2 + \epsilon} + \frac{1}{2} \!\right) \!\right)\!,
\end{equation}
subsequently, contextual features are obtained by weighted aggregation of $V$ guided by the enhanced attention weights $A'$, fused with the input sequence, projected through a feed-forward network (FFN), reshaped into spatial feature maps, and aligned with the resolution of the original features. The same bidirectional interaction is symmetrically applied to the NIR branch. Finally, the enhanced RGB and NIR features are concatenated along the channel dimension and processed by a $1\times1$ convolution to produce the fused feature $M_{fused}$, which can be mathematically expressed as follows:
\begin{align}
    Z_{RGB} &= \mathrm{FFN}\left( V_{RGB} \cdot  A'_{RGB} + T_{RGB} \right), \\
    Z_{NIR} &= \mathrm{FFN}\left( V_{NIR} \cdot  A'_{NIR}  + T_{NIR} \right), \\
    Z_{RGB}' &= \mathrm{UP}\left( \mathrm{Re}(Z_{RGB}) \right) + D_{RGB}, \\
    Z_{NIR}' &= \mathrm{UP}\left( \mathrm{Re}(Z_{NIR}) \right) + D_{NIR}, \\
    M_{fused} &= \mathrm{Conv}_{1\times1}\left( \mathrm{Cat}[ Z_{RGB}', Z_{NIR}']\right),
\end{align}
where $\mathrm{Re}(\cdot)$ denotes reshaping the sequence into a spatial tensor for restoring the spatial structure.
\subsection{SMRM}
To collaboratively refine fused features during the decoding stage, suppress pseudo-changes caused by modal differences, and enhance the spatial consistency of changed regions, we propose a Saliency-aware Multisource Refinement Module (SMRM), whose architecture is illustrated in Fig.~\ref{fig_2}. By incorporating external semantic priors and internal difference cues, this module enables precise focusing on the changed regions and progressively optimizes the feature representations, thereby enhancing the accuracy and reliability of building change detection tasks.

Specifically, saliency masks $\text{mask}{t_1}$ and $\text{mask}{t_2}$ are first introduced as spatial priors, which are generated offline by a pre-trained RemoteSAM model. These masks provide category-agnostic, object-level location awareness without introducing additional supervision during network training. To ensure spatial alignment, the saliency masks are resized via bilinear interpolation to match the spatial resolution of the highest-level feature map $M_4$. The resized masks are then concatenated with $M_4$ along the channel dimension and fused through a $1\times1$ convolution, which enhances contextual responses corresponding to regions of interest while suppressing background interference. This process leverages the semantic priors to spatially recalibrate the fused features, which can be formulated as:
\begin{equation}
M_4' = \mathrm{Conv}_{1\times1}\!\left(
\mathrm{Cat}[M_4, \mathrm{UP}(\mathrm{mask}_{t_1}), \mathrm{UP}(\mathrm{mask}_{t_2})]
\right)\!,
\end{equation}
feature refinement is subsequently propagated in a top-down manner through the proposed Multisource Refinement Block (MRB), which takes four types of inputs: the fusion feature $M_i$, the RGB difference feature $D_{RGB}$, the NIR difference feature $D_{NIR}$, and the enhanced feature from the previous scale $M_{i+1}'$. 
Using the fused representation, change difference cues and high-level semantic guidance are introduced to progressively refine features representations and enforce cross-scale consistency across hierarchical decoding stages.

The difference features are first linearly fused to aggregate complementary change cues from the RGB and NIR modalities while preserving their individual contributions, thereby providing temporal change information and enhancing the spatial localization of changed regions. The aggregated features are then passed through a $3\times3$ convolution to compress the channels. To ensure spatial consistency, the upper-level features are adjusted in both channels and resolution via a $3\times3$ convolution followed by bilinear interpolation. The procedures for feature alignment and compression are formulated as follows:
\begin{align}
D_i^1 &= \mathrm{Conv}_{3\times3}\left( D_i^{RGB} + D_i^{NIR} \right), \\
    M_{i+1}^1 &= \mathrm{UP}\left( \mathrm{Conv}_{3\times3}(M_{i+1}') \right),
\end{align}
next, the Fusion-guided Modulation Unit (FMU) is introduced to process the fused feature $M_i$, which encodes joint multi-modal change information at the current decoding stage. Specifically, local spatial features are first extracted through a $3\times3$ convolution, followed by global average pooling to obtain global channel statistics. These statistics are then passed through two fully connected layers with ReLU activation to generate channel-wise modulation weights, which are further normalized via the Sigmoid function to produce the channel modulation vector $A_M$. The modulation process is formulated as follows:
\begin{equation}
    A_M = \sigma\left( \mathrm{FC}\left( \delta\left( \operatorname{FC}\left( \operatorname{GAP}\left( \mathrm{Conv}_{3\times3}(M_i) \right) \right) \right) \right) \right),
\end{equation}
where $\mathrm{GAP(\cdot)}$ denotes the global average pooling operation, and $\mathrm{FC(\cdot)}$ represents the Fully Connected layer.

Adaptive refinement of multisource features is achieved through a learnable gating-based fusion strategy, where the coefficient $\omega_i \in [0,1]$ adaptively controls the contributions of difference cues and high-level semantic context and performs residual fusion. The fused feature is further refined by a $3\times3$ convolution and then upsampled to be propagated to the next decoding stage, and the refinement and fusion process is formulated as follows:
\begin{equation}
M_i' = \mathrm{UP}\Big(
\mathrm{Conv}_{3\times3}\big(\!
\omega_i \times D_i^1 \!+ (1-\omega_i) \times M_{i+1}^1\! +\! M_i^{\mathrm{fused}}
\big)\!\Big).
\end{equation}

\section{Experiment}\label{Section5}
\subsection{Dataset Description}
In addition to the proposed LSMD dataset, we conducted experiments on the SMARS~\cite{A_SMARS} multi-modal dataset to evaluate the generalization ability of our model. Derived from the SMARS dataset, SMARS provides synthetic orthophotos and DSM that simulate the urban environments of Paris and Venice. The dataset includes two spatial resolutions (0.3 m and 0.5 m), with land cover categorized into five classes: buildings, streets, trees, lawns, and others. At 0.3 m resolution, images from both cities have dimensions of $5,600 \times 5,600$ pixels, while at 0.5 m resolution, the image sizes are $4,500 \times 3,560$ pixels for Paris and $5,600 \times 5,600$ pixels for Venice.

\subsection{Comparison Methods and Evaluation Metrics}
To verify the effectiveness of the proposed method in utilizing multi-modal information, comprehensive quantitative comparisons are conducted with nine state-of-the-art methods on both the LSMD and SMARS datasets. These representative baselines include DCILNet~\cite{A_DCILNet}, S2CD~\cite{A_SSCD}, RHighNet~\cite{A_RHighNet}, ConvFormer-CD~\cite{A_ConvFormer}, HRMNet~\cite{A_HRMNet}, RFANet~\cite{A_RFANet}, CSI-Net~\cite{A_CSINet}, DGMA2-Net~\cite{A_DGMA}, and A2Net~\cite{A_A2Net}. To ensure fairness and comprehensiveness, we evaluate all baseline methods on the LSMD dataset under three input configurations: RGB-only, NIR-only, and RGB+NIR. For the baseline methods under the RGB+NIR configuration, we employ a channel concatenation strategy to combine RGB and NIR into a four-channel input. Similarly, on the SMARS dataset, the RGB and DSM data are combined into a four-channel input via channel concatenation and evaluated at both spatial resolutions. Performance is quantitatively evaluated using five widely adopted metrics, including Recall, Precision, F1-score (F1), Intersection over Union (IoU), and Kappa, with IoU and F1 as the primary indicators of overall change detection accuracy.

\subsection{Experimental Settings}
The proposed MSCNet is implemented using the PyTorch framework and trained and tested on a single NVIDIA GeForce RTX 3090 GPU. During training, the backbone network is initialized with ImageNet-pretrained MobileNetV2 weights to extract multi-scale feature representations. The model is optimized using the Adam optimizer with parameters set to $\beta_1 = 0.9$, $\beta_2 = 0.99$, and a weight decay of $1 \times 10^{-4}$. The initial learning rate is set to $5 \times 10^{-4}$ with a polynomial decay strategy. In addition, a linear warm-up strategy is applied during the first 200 iterations. The batch size is 16, and training runs for 40,000 iterations. The model is periodically evaluated on the validation set, and the checkpoint with the highest F1 is used for quantitative evaluation on the test dataset.

\subsection{Quantitative results}
\begin{table*}[htbp]
\centering
\caption{Quantitative comparison on the LSMD dataset using different input types. The best results are highlighted in \first{red}, the second-best results are \second{blue}, and the third-best results are \third{orange}.}
\label{table_2}
\resizebox{\textwidth}{!}{%
\begin{tabular}{c|c|cccccccccc}
\toprule
\multicolumn{2}{c|}{Datasets} & DCILNet & S2CD & RHighNet & ConvFormer-CD & HRMNet & RFANet & CSI-Net & DGMA2-Net & A2Net & MSCNet (ours)\\
\midrule
\multicolumn{2}{c|}{FLOPs(G)} & 41.47 & 9.55 & 66.51 & 5.14 & 12.17 & 3.16 & 367.25 & 18.10 & 3.05 & 4.49 \\
\midrule
\multicolumn{2}{c|}{Params(M)} & 26.80 & 3.26 & 96.81 & 37.72 & 13.46 & 2.86 & 62.18 & 37.10 & 3.78 & 6.40 \\
\midrule
\multirow{5}{*}{RGB} 
 & Kappa & 71.62 & 73.52 & 74.04 & 73.16 & 76.18 & 77.88 & 75.50 & \third{78.16} & \second{79.83} & \first{80.26} \\
 & IoU      & 56.96 & 59.23 & 59.91 & 58.85 & 62.62 & 64.82 & 61.72 & \third{65.20} & \second{67.40} & \first{67.98} \\
 & F1       & 72.58 & 74.40 & 74.93 & 74.10 & 77.01 & 78.66 & 76.33 & \third{78.93} & \second{80.52} & \first{80.94} \\
 & Recall      & 68.29 & 68.86 & 71.48 & 71.24 & 74.85 & 76.49 & 72.78 & \first{77.95} & \third{77.04} & \second{77.45} \\
 & Precision      & 77.45 & 80.90 & 78.73 & 77.19 & 79.31 & \third{80.95} & 80.26 & 79.95 & \second{84.34} & \first{84.76} \\
\midrule
\multirow{5}{*}{NIR} 
 & Kappa & 66.34 & 66.14 & 71.80 & 71.00 & 74.10 & 71.89 & 73.86 & \third{75.10} & \second{77.82} & \first{78.34} \\
 & IoU      & 50.87 & 50.51 & 57.12 & 56.21 & 59.98 & 57.25 & 59.65 & \third{61.23} & \second{64.72} & \first{65.41} \\
 & F1       & 67.43 & 67.12 & 72.70 & 71.97 & 74.98 & 72.81 & 74.72 & \third{75.96} & \second{78.58} & \first{79.09} \\
 & Recall      & 61.20 & 56.34 & 65.59 & 66.57 & 71.56 & 66.97 & 68.92 & \third{72.91} & \second{75.23} & \first{76.15} \\
 & Precision      & 75.08 & \first{83.01} & 81.55 & 78.32 & 78.76 & 79.77 & 81.59 & 79.27 & \third{82.25} & \second{82.26} \\
\midrule
\multirow{5}{*}{RGB+NIR} 
 & Kappa & 73.26 & 73.59 & 74.89 & 77.22 & 76.71 & 69.73 & \third{78.89} & 78.13 & \second{79.91} & \first{80.75} \\
 & IoU      & 58.91 & 59.29 & 60.96 & 63.94 & 63.29 & 54.70 & \third{66.11} & 65.19 & \second{67.51} & \first{68.67} \\
 & F1       & 74.15 & 74.44 & 75.74 & 78.00 & 77.52 & 70.72 & \third{79.60} & 78.92 & \second{80.60} & \first{81.42} \\
 & Recall      & 68.79 & 67.73 & 71.73 & 74.27 & 74.73 & 64.76 & 74.56 & \first{79.59} & \third{77.94} & \second{79.13} \\
 & Precision      & 80.40 & 82.63 & 80.23 & 82.14 & 80.52 & 77.89 & \first{85.38} & 78.27 & \third{83.45} & \second{83.86} \\
\bottomrule
\end{tabular}%
}
\end{table*}

\begin{table*}[htbp]
\centering
\caption{Quantitative comparison on SMARS datasets with different resolutions. The best results are highlighted in \first{red}, the second-best results are \second{blue}, and the third-best results are \third{orange}.}
\label{table_3}
\resizebox{\textwidth}{!}{%
\begin{tabular}{c|c|cccccccccc}
\toprule
\multicolumn{2}{c|}{\textbf{Datasets}} & DCILNet & S2CD & RHighNet & ConvFormer-CD & HRMNet & RFANet & CSI-Net & DGMA2-Net & A2Net & MSCNet (ours) \\ 
\midrule
\multirow{5}{*}{SMARS(0.3)}
 & Kappa   & 94.32 & 93.49 & 91.95 & 94.72 & \second{97.13} & 96.92 & 95.16 & 97.03 & \third{97.08} & \first{97.36} \\
 & IoU     & 91.26 & 90.02 & 87.81 & 91.85 & \second{95.48} & 95.17 & 92.50 & 95.33 & \third{95.41} & \first{95.84} \\
 & F1      & 95.43 & 94.75 & 93.51 & 95.75 & \second{97.69} & 97.53 & 96.11 & 97.61 & \third{97.65} & \first{97.88} \\
 & Recall  & 95.45 & 93.48 & 92.67 & 95.50 & 97.45 & \second{97.63} & 96.02 & \third{97.58} & 97.51 & \first{97.85} \\
 & Precision & 95.41 & 96.05 & 94.36 & 96.01 & \first{97.92} & 97.43 & 96.20 & 97.64 & \third{97.80} & \second{97.91} \\ 
\midrule
\multirow{5}{*}{SMARS(0.5)}
 & Kappa   & 91.63 & 89.28 & 88.71 & 92.00 & \second{95.26} & 94.91 & 94.78 & 94.86 & \third{94.92} & \first{95.64} \\
 & IoU     & 87.23 & 83.82 & 83.12 & 87.78 & \second{92.56} & 92.05 & 91.84 & 91.98 & \third{92.06} & \first{93.14} \\
 & F1      & 93.18 & 91.20 & 90.78 & 93.49 & \second{96.14} & 95.86 & 95.75 & 95.82 & \third{95.87} & \first{96.45} \\
 & Recall  & 92.99 & 88.00 & 90.00 & 93.19 & 96.10 & 96.24 & 95.89 & \second{96.48} & \third{96.33} & \first{96.50} \\
 & Precision & 93.37 & 94.64 & 91.58 & 93.80 & \second{96.18} & 95.48 & \third{95.61} & 95.17 & 95.41 & \first{96.39} \\ 
\bottomrule
\end{tabular}%
}
\end{table*}

To quantitatively evaluate the effectiveness of MSCNet, we conducted experiments on two datasets and compared it with nine change detection methods. In particular, experiments on the LSMD dataset were conducted under three input configurations (RGB-only, NIR-only, and RGB+NIR) to validate the effectiveness of multi-modal information and deep fusion mechanisms. The quantitative results on the LSMD dataset are summarized in Table~\ref{table_2}.

\subsubsection{Single-modal Baseline Performance}
Under single-modal input settings, various models exhibit significant feature extraction preferences and adaptability to modalities.

\textbf{RGB-only setting:} Under the RGB-only configuration, MSCNet achieves the best performance across most evaluation metrics, with Kappa, IoU, F1, Recall, and Precision reaching 80.26\%, 67.98\%, 80.94\%, 77.45\%, and 84.76\%, respectively; although the Recall is slightly lower than that of DGMA2-Net, it remains a highly competitive second place. The comprehensive lead of MSCNet across these indicators demonstrates that even with only visible light information, the model can effectively capture subtle changes through the precise feature discrimination capability.

\textbf{NIR-only setting:} In the NIR-only configuration, most methods experience a significant decline in performance due to the lack of rich texture and color information. For instance, compared to the performance with RGB inputs, the F1 and IoU of S2CD decrease by 7.28\% and 8.72\%, respectively. In contrast, MSCNet demonstrates exceptional robustness, with the F1 and IoU decreasing by only 1.85\% and 2.57\%. Notably, MSCNet achieves the best results across Kappa, IoU, F1, and Recall, while the Precision ranks second only to S2CD. This confirms the superior capability of the model to extract discriminative features even from the near-infrared single modality.

\subsubsection{Multi-modal Fusion Performance}
When RGB and NIR data are jointly employed, most methods show performance improvements, validating the complementary nature of multi-modal information. Notably, MSCNet achieves the best overall performance.

\textbf{Optimal Comprehensive Performance:} 
User 21:15
Under the RGB+NIR configuration, MSCNet achieves Kappa, IoU, F1, Recall, and Precision of 80.75\%, 68.67\%, 81.42\%, 79.13\%, and 83.86\%, respectively. Compared with single-modal inputs, all evaluation metrics exhibit further improvements after fusion, confirming that the model realizes efficient cross-modal feature integration to better handle illumination variations and pseudo-change interferences.

\textbf{Necessity of Deep Fusion Mechanisms:} It is worth noting that not all methods benefit from multi-modal inputs. For instance, RFANet suffers from performance degradation after the introduction of NIR data, with the IoU and F1 decreasing by 10.12\% and 7.94\%, respectively. This result suggests that simple channel concatenation fusion methods cannot model cross-modal interactions effectively and may even introduce noise. In contrast, MSCNet consistently outperforms other state-of-the-art methods under the multi-modal setting, confirming the necessity of designing dedicated deep fusion mechanisms.

\begin{figure*}[!t]  
    \centering
    \includegraphics[width=\textwidth]{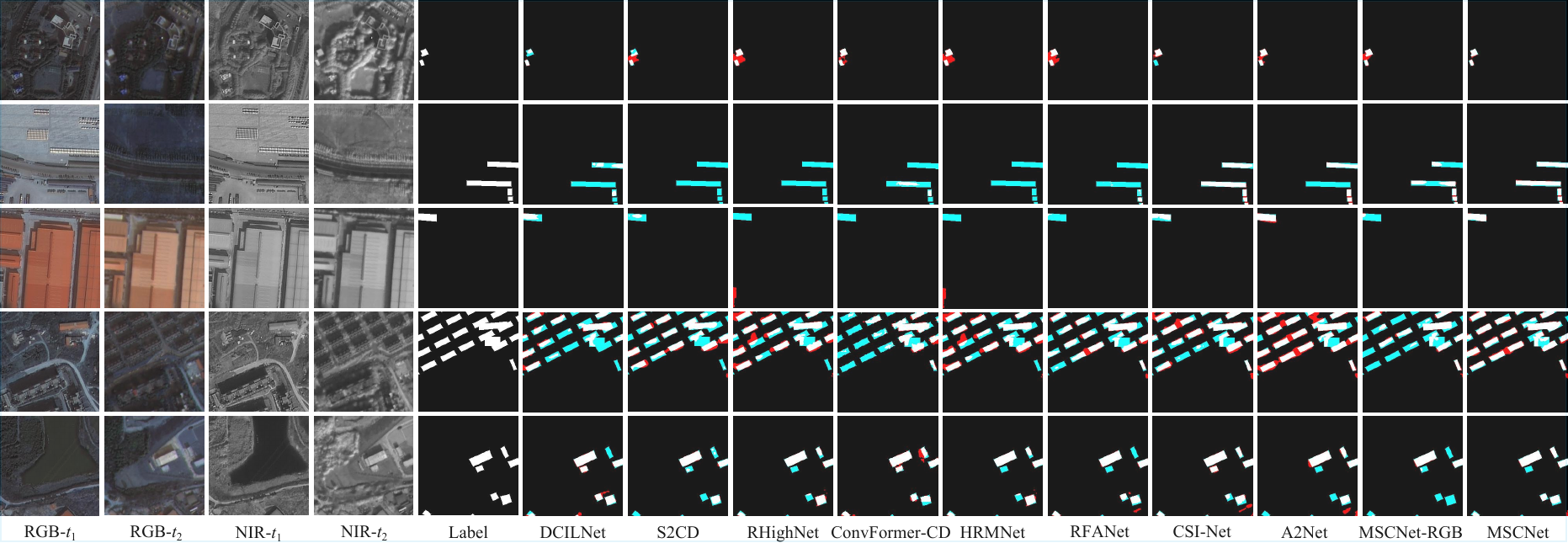} 
    \caption{Qualitative visualizations of different change detection approaches on the LSMD dataset under RGB-only configuration, with MSCNet using multi-modal (RGB+NIR) information for direct comparison. The rendered colors denote TP (white), FP (red), TN (black), and FN (blue).}
    \label{fig_6}
\end{figure*}

\textbf{Performance on SMARS Dataset: } To evaluate the robustness of MSCNet across diverse multi-modal scenarios and varying spatial resolutions, comparative experiments were conducted on the SMARS dataset at 0.3m and 0.5m resolutions, as reported in Table~\ref{table_3}. MSCNet achieves the best overall performance at both resolution levels, leading in most evaluation metrics, particularly the critical IoU and F1 scores. Notably, while advanced models like HRMNet and A2Net deliver competitive results at 0.3m resolution, their performance suffers more significant degradation as the resolution decreases to 0.5m. In contrast, MSCNet maintains a robust F1 score of 96.45\% and an IoU of 93.14\% in the more challenging 0.5m setting, exhibiting superior numerical stability compared to other approaches. These experimental results provide compelling evidence that our multi-modal feature fusion strategy effectively captures complementary information, ensuring the model remains reliable across different spatial scales.

\textbf{Parameter Size and Computational Complexity: } In addition to detection accuracy, computational efficiency is a crucial factor for practical RSCD applications. As shown in Table~\ref{table_2}, some methods achieve competitive accuracy at the cost of significantly increased computational complexity. In contrast, MSCNet contains only 6.40 M parameters and 4.49 G FLOPs, which are substantially lower than those of most structurally complex or Transformer-based baselines, such as CSI-Net, RHighNet, and DGMA2-Net. Nevertheless, MSCNet still attains superior or competitive performance across all evaluation metrics. These results demonstrate that MSCNet achieves a favorable balance between accuracy and efficiency for multi-modal change detection.

\subsection{Qualitative analysis}

\begin{figure*}[!t]  
    \centering
    \includegraphics[width=\textwidth]{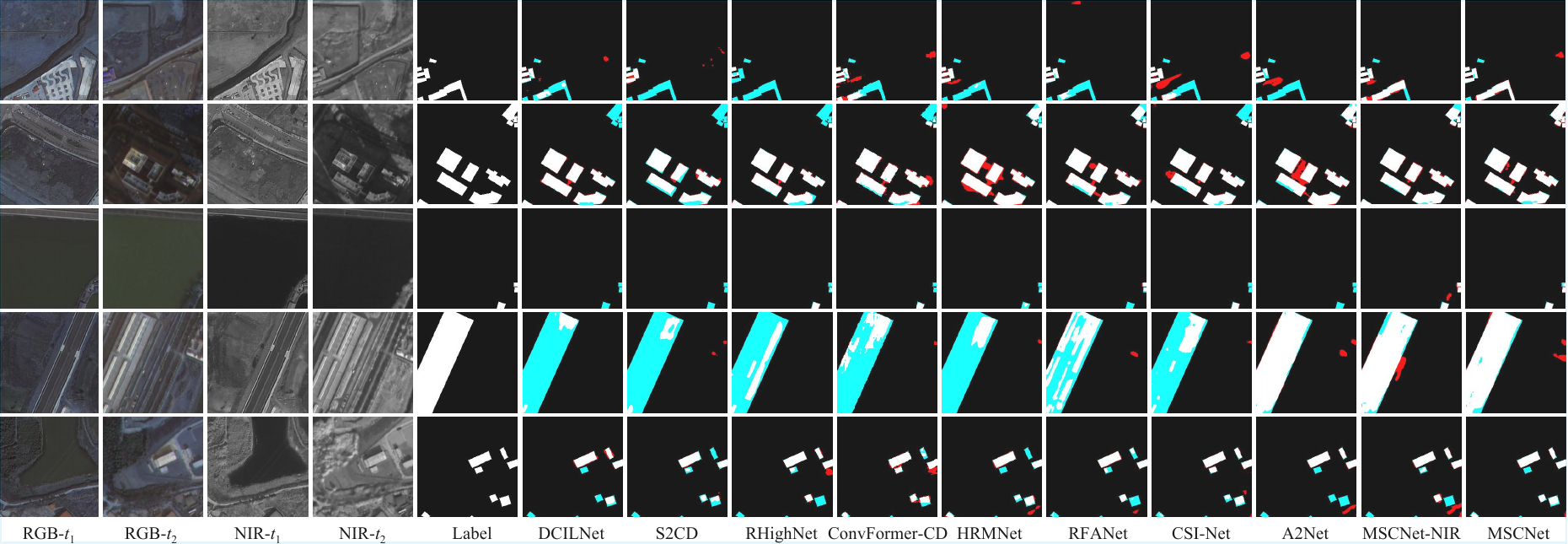} 
    \caption{Qualitative visualizations of different change detection approaches on the LSMD dataset under NIR-only configuration, with MSCNet using multi-modal (RGB+NIR) information for direct comparison. The rendered colors denote TP (white), FP (red), TN (black), and FN (blue).}
    \label{fig_7}
\end{figure*}

\begin{figure*}[!t]  
    \centering
    \includegraphics[width=\textwidth]{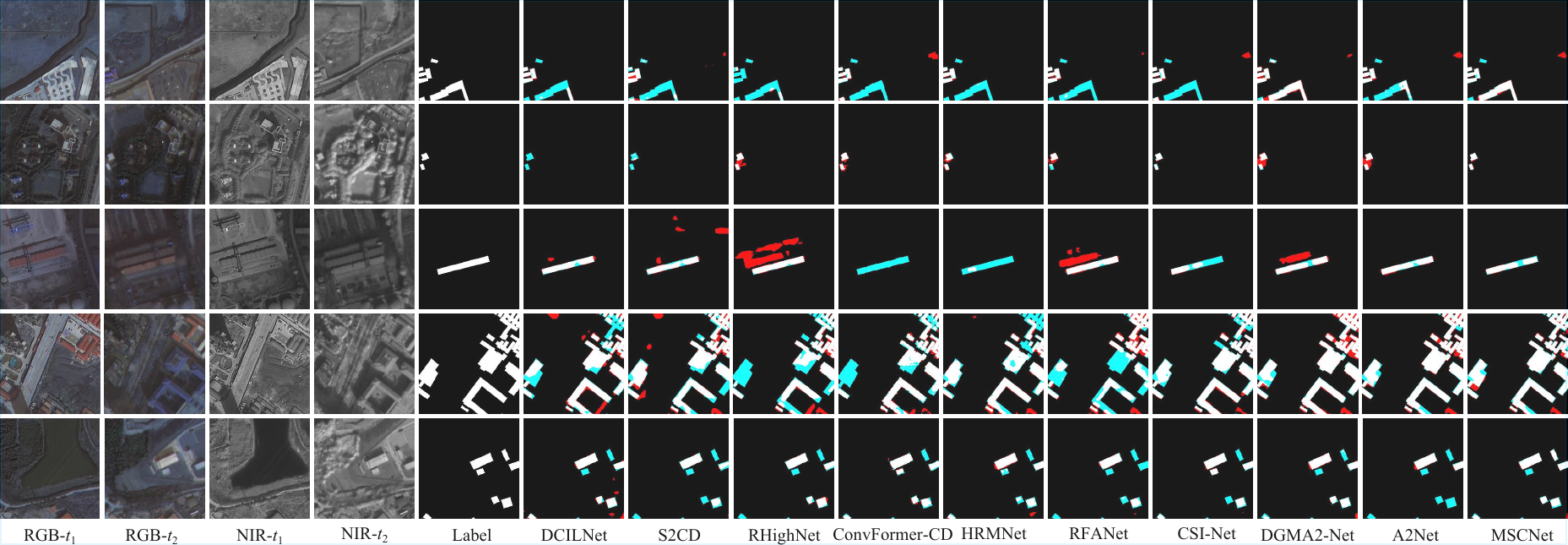} 
    \caption{Qualitative visualizations of different change detection approaches on the LSMD dataset under multi-modal (RGB+NIR) configuration. The rendered colors denote TP (white), FP (red), TN (black), and FN (blue).}
    \label{fig_8}
\end{figure*}

\begin{figure*}[!t]  
    \centering
    \includegraphics[width=\textwidth]{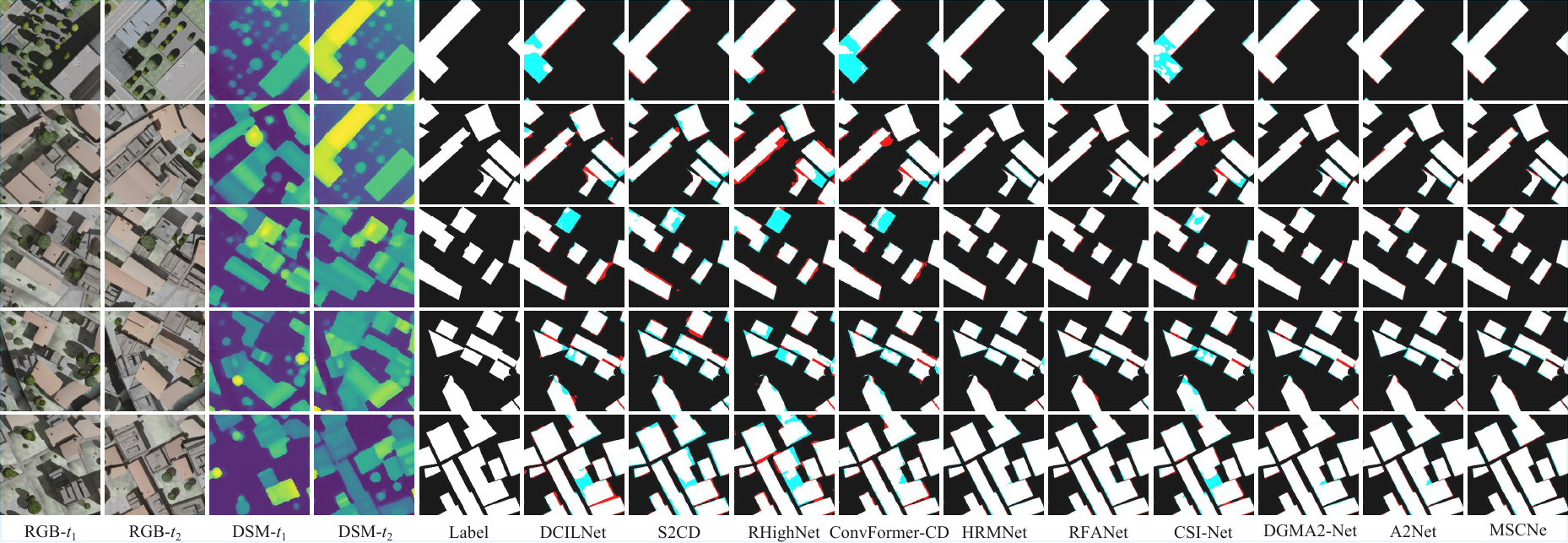} 
    \caption{Qualitative visualizations of different change detection approaches on the SMARS dataset at 0.3 m spatial resolution. The rendered colors denote TP (white), FP (red), TN (black), and FN (blue).}
    \label{fig_9}
\end{figure*}
\begin{figure*}  
    \centering
    \includegraphics[width=\textwidth]{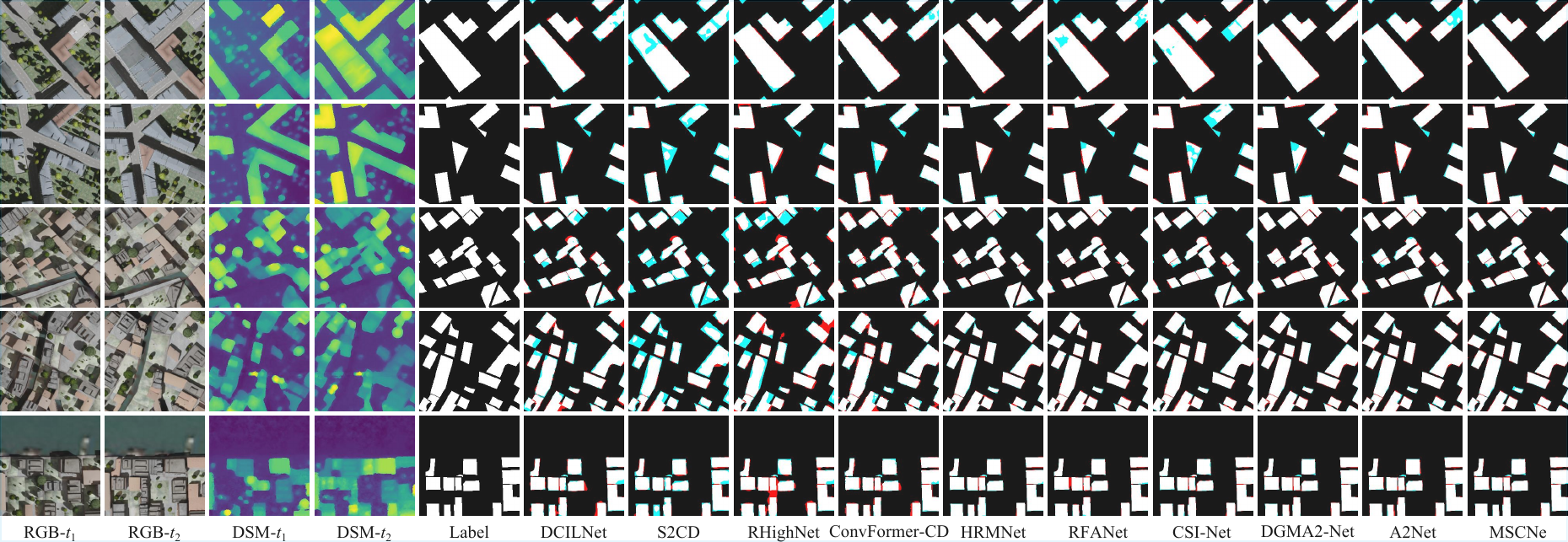} 
    \caption{Qualitative visualizations of different change detection approaches on the SMARS dataset at 0.5 m spatial resolution. The rendered colors denote TP (white), FP (red), TN (black), and FN (blue).}
    \label{fig_10}
\end{figure*}

To intuitively evaluate and compare the detection performance of different change detection methods, we visualize the results on both the LSMD dataset (under three input configurations: RGB-only, NIR-only, and RGB+NIR) and the SMARS dataset. The corresponding qualitative results for LSMD are shown in Figs.~\ref{fig_6}-\ref{fig_8}, while those for SMARS are shown in Figs.~\ref{fig_9}-\ref{fig_10}. In the visualizations, different colors indicate different detection outcomes: white represents true positives (TP), red represents false positives (FP), black represents true negatives (TN), and blue represents false negatives (FN), highlighting areas where changes were missed by the models.

Figs.~\ref{fig_6}-\ref{fig_8} present the qualitative results on the LSMD dataset. First, the proposed method is able to more accurately localize real change regions. For instance, as illustrated in the first row of Fig.~\ref{fig_6}, the third row of Fig.~\ref{fig_7}, and the second row of Fig.~\ref{fig_8}, many competing methods fail to extract complete features when dealing with small-scale building changes, leading to apparent missed detections or false alarms. Although the proposed method maintains reasonable performance under single-modality inputs, its capability is inevitably constrained in extreme fine-grained scenarios (e.g., the fifth row of Figs.~\ref{fig_6}–\ref{fig_8}) due to the limited physical information of a single spectral modality. In contrast, the RGB+NIR configuration effectively compensates for such perceptual blind spots and yields clear advantages in detection completeness and boundary continuity. This shows that our model goes beyond simple feature concatenation by using a carefully designed fusion strategy, which effectively recovers fine-grained target structures and thus enables more complete detection of changes while representing structural details more accurately.

Second, the proposed method is capable of reliably identifying change regions in environments that are visually similar to buildings. As shown in the second and third rows of Fig.~\ref{fig_6} and the third row of Fig.~\ref{fig_8}, visually similar background regions often confuse existing CD methods, leading to misclassification between true changes and background clutter. By contrast, our method effectively distinguishes spectral discrepancies from genuine changes, robustly suppressing such interference to produce much cleaner and more accurate~detection results.

Moreover, the visualization results clearly demonstrate the complementary value of RGB and NIR information. Change regions that are difficult to recognize in the RGB modality can be effectively compensated by incorporating NIR information. As illustrated in the fourth row of Fig.~\ref{fig_6}, insufficient illumination in RGB images causes blurred building boundaries, leading to widespread missed detections by RGB-only methods. After introducing NIR information, the model leverages the sensitivity of the NIR band to building materials to successfully recover the missing contours. Conversely, for change regions with ambiguous boundaries in the NIR modality, RGB color information provides complementary cues for boundary refinement. As shown in the bottom-right area of the first row in Fig.~\ref{fig_7}, regions with unclear textures in NIR benefit from RGB color information, resulting in boundaries that better align with the ground truth and thereby significantly reducing both false positives and false negatives across the entire image.

Furthermore, qualitative results on the SMARS dataset with two different spatial resolutions (0.3 m and 0.5 m) are presented in Figs.~\ref{fig_9}-\ref{fig_10}. Compared with competing methods such as DCILNet and S2CD, which exhibit severe fragmented missed detections and background false-alarm noise in complex urban scenes, the proposed MSCNet, benefiting from the deep fusion of RGB imagery and DSM elevation information, effectively mitigates the influence of shadows and texture interference. The resulting detection maps not only substantially reduce false positives and false negatives, but also achieve superior visual quality in terms of internal building completeness and boundary sharpness. 

\subsection{Ablation Study}
\begin{table}[htbp]
\centering
\small
\caption{Ablation study of different modules on the LSMD dataset.}
\label{table_4}
\resizebox{\columnwidth}{!}{%
\begin{tabular}{c|l|cc}
\toprule
\multirow{2}{*}{Methods} & \multirow{2}{*}{Variants} & \multicolumn{2}{c}{LSMD} \\
\cmidrule(l){3-4}
 &   & IoU & F1 \\
\midrule
(a) & w/o & 64.48 & 78.41 \\
(b) & Ours & 68.67 & 81.42 \\
\midrule
\multicolumn{4}{l}{(1) Neighborhood Context Enhancement Module (NCEM)} \\
\midrule
(c) & only NCEM & 65.48 & 79.14 \\
(d) & w/o NCEM & 65.93 & 79.47 \\
(e) & w/o NCEM + FRM & 67.23 & 80.41 \\
\midrule
\multicolumn{4}{l}{(2) Cross-modal Alignment and Interaction Module (CAIM)} \\
\midrule
(f) & only CAIM & 65.76 & 79.35 \\
(g) & w/o CAIM & 67.85 & 80.84 \\
(h) & w/o BDCA & 67.02 & 80.26 \\
(i) & w/o Variance Weighting & 67.62 & 80.69 \\
(j) & w/o CCAF & 68.16 & 81.07 \\
\midrule
\multicolumn{4}{l}{(3) Saliency-aware Multisource Refinement Module (SMRM)} \\
\midrule
(k) & only SMRM & 65.92 & 79.46 \\
(l) & w/o SMRM & 67.73 & 80.76 \\
(m) & w/o RemoteSAM mask & 68.09 & 81.01 \\
(n) & w/o diff-guided \& cross-layer fusion & 65.16 & 78.90 \\
\bottomrule
\end{tabular}%
}
\end{table}

\begin{figure*}  
    \centering
    \includegraphics[width=\textwidth]{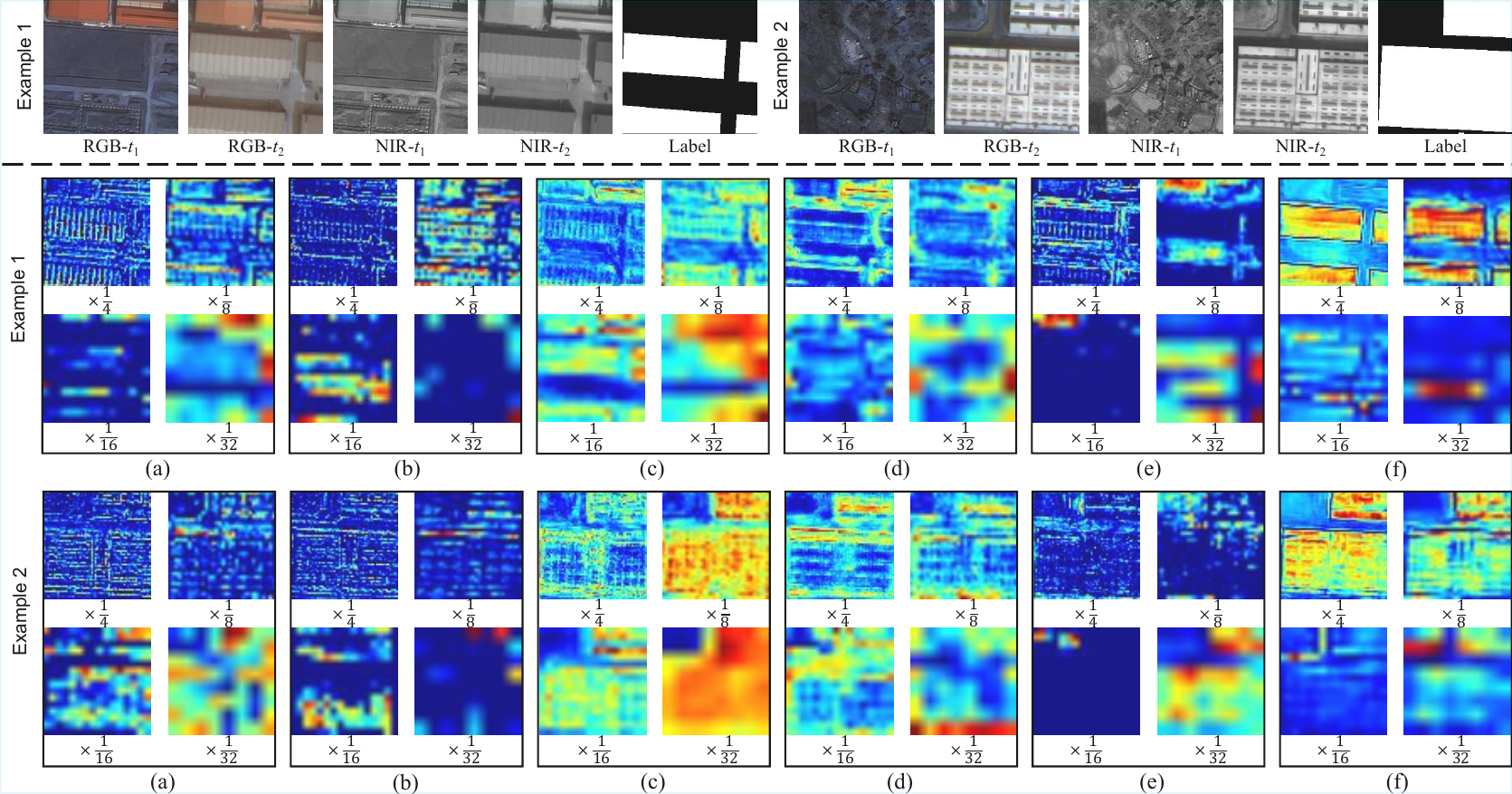} 
    \caption{Heatmap visualizations of the proposed modules. (a) and (b) show the RGB${t_2}$ and NIR${t_2}$ feature maps extracted from the four stages of the backbone network, respectively. (c) and (d) present the RGB${t_2}$ and NIR${t_2}$ feature maps after applying the NCEM. (e) shows the feature maps after the CAIM. (f) illustrates the feature maps refined by the SMRM.}
    \label{fig_11}
\end{figure*}
To validate the effectiveness of the proposed modules, comprehensive ablation studies are conducted on the LSMD dataset, and the quantitative results illustrating the contributions of different components are reported in Table~\ref{table_4}. Specifically, (a) denotes the baseline model without NCEM, CAIM, and SMRM; (b) represents the full model incorporating all proposed modules; (c), (f), and (k) correspond to models equipped with only a single module; while the remaining configurations show the performance obtained by either removing the corresponding module or replacing existing modules within the full model.

\textbf{Effectiveness of NCEM: } By comparing configurations (a)–(e), the following observations can be made. Introducing NCEM alone (c) improves the IoU and F1 scores to 65.48\% and 79.14\%, corresponding to gains of 1.00\% and 0.73\% over the baseline (a). In contrast, removing NCEM from the full model (b) results in noticeable performance degradation (d), with IoU and F1 dropping by 2.74\% and 1.95\%, respectively, highlighting the critical role of NCEM in the overall architecture. Furthermore, we compared NCEM with the feature reinforcement module FRM (e)~\cite{A_RFANet}. While FRM (e) outperforms the variant without enhancement (d) due to its dense multi-scale interaction, our NCEM (b) achieves a 1.44\% higher IoU than FRM. This confirms the effectiveness of the proposed strategy of selectively extracting and propagating key features between adjacent layers, which enhances spatial details while effectively avoiding the information redundancy caused by over-fusion.

\textbf{Effectiveness of CAIM: } By comparing configurations (a), (b), and (g)–(j), the following observations can be made. Introducing CAIM alone (f) improves IoU and F1 by 1.28\% and 0.94\% over the baseline (a), indicating that CAIM contributes to effective RGB–NIR feature fusion. In contrast, removing CAIM from the full model (b) results in performance degradation (g), with IoU and F1 decreasing by 0.82\% and 0.58\%, respectively. Moreover, replacing the BDCA with simple feature concatenation (h) leads to a noticeable performance drop, highlighting the importance of explicitly modeling cross-modal interactions. Removing key BDCA submodules, including variance weighting and CCAF, in (i) and (j) further degrades performance, indicating that the coordinated design of CAIM components is critical for robust multi-modal feature fusion.

\textbf{Effectiveness of SMRM: }By comparing configurations (a), (b), and (k)-(n), the following observations can be made. The model equipped only with the SMRM (k) achieves improvements of 1.44\% in IoU and 1.05\% in F1 over the baseline, indicating that multisource feature refinement under the guidance of semantic masks contributes to more effective modeling of fine-grained change patterns. When SMRM is removed (l), IoU and F1 decrease by 0.94\% and 0.66\%, suggesting that SMRM contributes to improved semantic change perception and cross-scale feature consistency. Furthermore, removing the semantic priors provided by RemoteSAM, or excluding the auxiliary information from multi-modal difference features and features from the previous scale, results in performance degradation to varying degrees. These results suggest that the joint incorporation of external semantic priors, multi-modal difference cues, and cross-layer feature refinement contributes to improved sensitivity to complex and subtle changes.

To further verify the effectiveness of the proposed modules, we visualize the feature responses from the last four stages of the network and present the corresponding heatmaps after sequentially applying each module, as shown in Fig.~\ref{fig_11}. It can be observed that the proposed modules progressively enhance the model’s responses to changed regions while effectively suppressing background interference.

\section{Conclusion}
This article investigates multi-modal building change detection by integrating the distinct physical attributes of RGB and NIR domains to address spectral ambiguity in single-modality imagery. We construct a high-resolution bi-temporal benchmark dataset, LSMD, which provides authentic RGB and NIR data and focuses on two practical challenges in real-world scenarios: small changes in large-scale scenes and interference from vegetation backgrounds. Based on this, we propose MSCNet, leveraging NCEM to enhance local context, CAIM for deep cross-modal feature interactions, and SMRM to refine fused features. Extensive experiments show that MSCNet consistently outperforms state-of-the-art methods, confirming its effectiveness for fine-grained building change detection. For future work, we aim to optimize multi-modal feature fusion to fully exploit complementary information and enhance sensitivity to subtle structural changes. We also plan to explore lightweight architectures and dynamic feature weighting strategies to enhance efficiency, and adapt the framework for multi-class fine-grained building change detection.

\section*{Acknowledgements}
This work was supported in part by the NSFC Key Project of Joint Fund for Enterprise Innovation and Development under Grant U24A20342, and in part by the National Natural Science Foundation of China under Grant 62576006 and 61976004.


\bibliographystyle{elsarticle-num-names}
\bibliography{LSMD}

@Article{A_LEVIR,
AUTHOR = {Chen, Hao and Shi, Zhenwei},
TITLE = {A Spatial-Temporal Attention-Based Method and a New Dataset for Remote Sensing Image Change Detection},
JOURNAL = {Remote Sensing},
VOLUME = {12},
YEAR = {2020},
NUMBER = {10},
ARTICLE-NUMBER = {1662},
URL = {https://www.mdpi.com/2072-4292/12/10/1662},
ISSN = {2072-4292},
DOI = {10.3390/rs12101662}
}

@ARTICLE{A_WHU,
  author={Ji, Shunping and Wei, Shiqing and Lu, Meng},
  journal={IEEE Transactions on Geoscience and Remote Sensing}, 
  title={Fully Convolutional Networks for Multisource Building Extraction From an Open Aerial and Satellite Imagery Data Set}, 
  year={2019},
  volume={57},
  number={1},
  pages={574-586},
  doi={10.1109/TGRS.2018.2858817}}

@article{A_CDD,
  title={Change detection in remote sensing images using conditional adversarial networks},
  author={Lebedev, MA and Vizilter, Yu V and Vygolov, OV and Knyaz, Vladimir A and Rubis, A Yu},
  journal={The International Archives of the Photogrammetry, Remote Sensing and Spatial Information Sciences},
  volume={42},
  pages={565--571},
  year={2018},
  publisher={Copernicus GmbH}
}

@ARTICLE{A_SYSU,
  author={Shi, Qian and Liu, Mengxi and Li, Shengchen and Liu, Xiaoping and Wang, Fei and Zhang, Liangpei},
  journal={IEEE Transactions on Geoscience and Remote Sensing}, 
  title={A Deeply Supervised Attention Metric-Based Network and an Open Aerial Image Dataset for Remote Sensing Change Detection}, 
  year={2022},
  volume={60},
  number={},
  pages={1-16},
  doi={10.1109/TGRS.2021.3085870}}

@article{A_xiongan,
  title={{HeteCD}: Feature Consistency Alignment and difference mining for heterogeneous remote sensing image change detection},
  author={Jing, Wei and Bai, Haichen and Song, Binbin and Ni, Weiping and Wu, Junzheng and Wang, Qi},
  journal={ISPRS Journal of Photogrammetry and Remote Sensing},
  volume={223},
  pages={317--327},
  year={2025},
  publisher={Elsevier}
}

@article{A_MT-wuhan,
  title={A domain adaptation neural network for change detection with heterogeneous optical and SAR remote sensing images},
  author={Zhang, Chenxiao and Feng, Yukang and Hu, Lei and Tapete, Deodato and Pan, Li and Liang, Zheheng and Cigna, Francesca and Yue, Peng},
  journal={International Journal of Applied Earth Observation and Geoinformation},
  volume={109},
  pages={102769},
  year={2022},
  publisher={Elsevier}
}

@article{A_3DCD,
  title={Inferring 3D change detection from bitemporal optical images},
  author={Marsocci, Valerio and Coletta, Virginia and Ravanelli, Roberta and Scardapane, Simone and Crespi, Mattia},
  journal={ISPRS Journal of Photogrammetry and Remote Sensing},
  volume={196},
  pages={325--339},
  year={2023},
  publisher={Elsevier}
}

@article{A_SMARS,
  title={A 2D/3D multimodal data simulation approach with applications on urban semantic segmentation, building extraction and change detection},
  author={Reyes, Mario Fuentes and Xie, Yuxing and Yuan, Xiangtian and d’Angelo, Pablo and Kurz, Franz and Cerra, Daniele and Tian, Jiaojiao},
  journal={ISPRS Journal of Photogrammetry and Remote Sensing},
  volume={205},
  pages={74--97},
  year={2023},
  publisher={Elsevier}
}

@article{A_OSCD,
  title={Fusing multi-modal data for supervised change detection},
  author={Ebel, Patrick and Saha, Sudipan and Zhu, Xiao Xiang},
  journal={The international archives of the photogrammetry, remote sensing and spatial information sciences},
  volume={43},
  pages={243--249},
  year={2021},
  publisher={Copernicus GmbH}
}

@misc{A_RemoteSAM,
      title={{RemoteSAM}: Towards Segment Anything for Earth Observation}, 
      author={Liang Yao and Fan Liu and Delong Chen and Chuanyi Zhang and Yijun Wang and Ziyun Chen and Wei Xu and Shimin Di and Yuhui Zheng},
      year={2025},
      eprint={2505.18022},
      archivePrefix={arXiv},
      primaryClass={cs.CV},
      url={https://arxiv.org/abs/2505.18022}, 
}

@article{A_GLCD,
  title={{GLCD-DA}: Change detection from optical and SAR imagery using a Global-Local network with diversified attention},
  author={Li, Jie and Wu, Meiru and Lin, Liupeng and Yuan, Qiangqiang and Shen, Huanfeng},
  journal={ISPRS Journal of Photogrammetry and Remote Sensing},
  volume={226},
  pages={396--414},
  year={2025},
  publisher={Elsevier}
}

@INPROCEEDINGS{A_HGN,
  author={Lyu, Jialin and Fu, Yimin and Liu, Zhunga},
  booktitle={IGARSS 2025 - 2025 IEEE International Geoscience and Remote Sensing Symposium}, 
  title={Hierarchical Gated Network for Multimodal Remote Sensing Imagery Classification with Limited Data}, 
  year={2025},
  volume={},
  number={},
  pages={2672-2676},
  doi={10.1109/IGARSS55030.2025.11243186}}

@INPROCEEDINGS{A_MCTUNet,
  author={Ren, Bin and Shi, Zewei and Fan, Hexiao and He, Chunhong},
  booktitle={2025 6th International Conference on Electronic Communication and Artificial Intelligence (ICECAI)}, 
  title={A Remote Sensing Image Segmentation Network Based on Multimodal Feature Fusion Enhancement}, 
  year={2025},
  volume={},
  number={},
  pages={319-322},
  doi={10.1109/ICECAI66283.2025.11170498}}

@INPROCEEDINGS{A_MDFNet,
  author={Wei, Tianyu and Chen, He and Wang, Jue and Liu, Wenchao},
  booktitle={2024 IEEE International Conference on Signal, Information and Data Processing (ICSIDP)}, 
  title={{MDFNet}: Multimodal Feature Decomposition and Fusion Network for Multimodal Remote Sensing Image Semantic Segmentation}, 
  year={2024},
  volume={},
  number={},
  pages={1-5},
  doi={10.1109/ICSIDP62679.2024.10868650}}

@INPROCEEDINGS{A_RCAM,
  author={Li, Haitao and Yang, Zongyu and Gu, Haiyan and Yang, Yi and Shen, Hengtong and Kong, Haozhu and Li, Honglin and Wang, Yang},
  booktitle={IGARSS 2025 - 2025 IEEE International Geoscience and Remote Sensing Symposium}, 
  title={Remote Sensing Image Change Detection Method Based on Segment Anything Model}, 
  year={2025},
  volume={},
  number={},
  pages={6629-6632},
  doi={10.1109/IGARSS55030.2025.11243280}}

@ARTICLE{A_FAEWNet,
  author={Li, Yun-Cheng and Lei, Sen and Zhao, Yi-Tao and Li, Heng-Chao and Li, Jun and Plaza, Antonio},
  journal={IEEE Transactions on Geoscience and Remote Sensing}, 
  title={{SAM}-Based Building Change Detection With Distribution-Aware Fourier Adaptation and Edge-Constrained Warping}, 
  year={2025},
  volume={63},
  number={},
  pages={1-14},
  doi={10.1109/TGRS.2025.3629110}}

@article{A_DCILNet,
  title={Dual-Branch Cross-Resolution Interaction Learning Network for Change Detection at Different Resolutions},
  author={Li, Jinghui and Shao, Feng and Meng, Xiangchao and Yang, Zhiwei},
  journal={IEEE Transactions on Geoscience and Remote Sensing},
  year={2024},
  publisher={IEEE}
}

@ARTICLE{A_SSCD,
  author={Wang, Leiquan and Fang, Ye and Li, Zhongwei and Wu, Chunlei and Xu, Mingming and Shao, Mingwen},
  journal={IEEE Transactions on Geoscience and Remote Sensing}, 
  title={{Summator–Subtractor Network}: Modeling Spatial and Channel Differences for Change Detection}, 
  year={2024},
  volume={62},
  number={},
  pages={1-12},
  doi={10.1109/TGRS.2024.3349638}}

@ARTICLE{A_RHighNet,
  author={Dong, Huihui and Du, Xinyu and Li, Zhijie and Ma, Zongfang and Wang, Yumeng and Zhu, Hao and Ma, Wenping},
  journal={IEEE Transactions on Geoscience and Remote Sensing}, 
  title={Relation-Aware High-Order Interaction Network for Remote Sensing Image Change Detection}, 
  year={2025},
  volume={63},
  number={},
  pages={1-16},
  doi={10.1109/TGRS.2025.3604400}}

@ARTICLE{A_ConvFormer,
  author={Yang, Feng and Li, Mengtao and Shu, Wenqiang and Qin, Anyong and Song, Tiecheng and Gao, Chenqiang and Xia, Gui-Song},
  journal={IEEE Transactions on Geoscience and Remote Sensing}, 
  title={{ConvFormer-CD}: Hybrid CNN–Transformer With Temporal Attention for Detecting Changes in Remote Sensing Imagery}, 
  year={2025},
  volume={63},
  number={},
  pages={1-15},
  doi={10.1109/TGRS.2025.3544651}}

@ARTICLE{A_HRMNet,
  author={Li, Zhenglai and Tang, Chang and Hu, Xingchen and Li, Ning and Xiang, Sen and Li, Chuankun and Li, Changdong and Liu, Xinwang},
  journal={IEEE Transactions on Geoscience and Remote Sensing}, 
  title={Boosting Remote Sensing Change Detection via Hard Region Mining}, 
  year={2025},
  volume={63},
  number={},
  pages={1-12},
  doi={10.1109/TGRS.2025.3598766}}

@article{A_RFANet,
  title={Robust feature aggregation network for lightweight and effective remote sensing image change detection},
  author={You, Zhi-Hui and Chen, Si-Bao and Wang, Jia-Xin and Luo, Bin},
  journal={ISPRS Journal of Photogrammetry and Remote Sensing},
  volume={215},
  pages={31--43},
  year={2024},
  publisher={Elsevier}
}

@ARTICLE{A_CSINet,
  author={Liu, Yunlong and Zhang, Feng and Zhang, Shanxin and Zhang, Kai and Sun, Jiande and Bruzzone, Lorenzo},
  journal={IEEE Transactions on Geoscience and Remote Sensing}, 
  title={Content-Guided Spatial–Spectral Integration Network for Change Detection in HR Remote Sensing Images}, 
  year={2024},
  volume={62},
  number={},
  pages={1-16},
  doi={10.1109/TGRS.2024.3352050}}

@ARTICLE{A_DGMA,
  author={Ying, Zilu and Tan, Zijun and Zhai, Yikui and Jia, Xudong and Li, Wenba and Zeng, Junying and Genovese, Angelo and Piuri, Vincenzo and Scotti, Fabio},
  journal={IEEE Transactions on Geoscience and Remote Sensing}, 
  title={{DGMA2-Net}: A Difference-Guided Multiscale Aggregation Attention Network for Remote Sensing Change Detection}, 
  year={2024},
  volume={62},
  number={},
  pages={1-16},
  doi={10.1109/TGRS.2024.3390206}}

@ARTICLE{A_A2Net,
  author={Li, Zhenglai and Tang, Chang and Liu, Xinwang and Zhang, Wei and Dou, Jie and Wang, Lizhe and Zomaya, Albert Y.},
  journal={IEEE Transactions on Geoscience and Remote Sensing}, 
  title={Lightweight Remote Sensing Change Detection With Progressive Feature Aggregation and Supervised Attention}, 
  year={2023},
  volume={61},
  number={},
  pages={1-12},
  doi={10.1109/TGRS.2023.3241436}}

@article{A_wang2024advances,
  title={Advances and challenges in deep learning-based change detection for remote sensing images: A review through various learning paradigms},
  author={Wang, Lukang and Zhang, Min and Gao, Xu and Shi, Wenzhong},
  journal={Remote Sensing},
  volume={16},
  number={5},
  pages={804},
  year={2024},
  publisher={MDPI}
}

@article{A_shafique2022deep,
  title={Deep learning-based change detection in remote sensing images: A review},
  author={Shafique, Ayesha and Cao, Guo and Khan, Zia and Asad, Muhammad and Aslam, Muhammad},
  journal={Remote Sensing},
  volume={14},
  number={4},
  pages={871},
  year={2022},
  publisher={MDPI}
}

@article{A_cheng2024change,
  title={Change detection methods for remote sensing in the last decade: A comprehensive review},
  author={Cheng, Guangliang and Huang, Yunmeng and Li, Xiangtai and Lyu, Shuchang and Xu, Zhaoyang and Zhao, Hongbo and Zhao, Qi and Xiang, Shiming},
  journal={Remote Sensing},
  volume={16},
  number={13},
  pages={2355},
  year={2024},
  publisher={MDPI}
}

@article{A_DSM,
  title={Building change detection based on 3D co-segmentation using satellite stereo imagery},
  author={Wang, Hao and Lv, Xiaolei and Zhang, Kaiyu and Guo, Bin},
  journal={Remote Sensing},
  volume={14},
  number={3},
  pages={628},
  year={2022},
  publisher={MDPI}
}

@article{A_SAR_juxian,
  title={Deep-learning for change detection using multi-modal fusion of remote sensing images: a review},
  author={Saidi, Souad and Idbraim, Soufiane and Karmoude, Younes and Masse, Antoine and Arbelo, Manuel},
  journal={Remote Sensing},
  volume={16},
  number={20},
  pages={3852},
  year={2024},
  publisher={MDPI}
}

@article{A_DSM_juxian,
  title={Enhancing Photogrammetric DSM Based on Multiscale and Domain-Invariant Semantic Feature Learning},
  author={Wang, Xuanqi and Jiang, Liting and Xiang, Yuming and Jiao, Niangang and Yang, Wen and Wang, Feng and You, Hongjian},
  journal={IEEE Journal of Selected Topics in Applied Earth Observations and Remote Sensing},
  volume={18},
  pages={28677--28694},
  year={2025},
  publisher={IEEE}
}

@article{A_SAR,
  title={Supervised change detection using prechange optical-SAR and postchange SAR data},
  author={Saha, Sudipan and Shahzad, Muhammad and Ebel, Patrick and Zhu, Xiao Xiang},
  journal={IEEE Journal of Selected Topics in Applied Earth Observations and Remote Sensing},
  volume={15},
  pages={8170--8178},
  year={2022},
  publisher={IEEE}
}

@INPROCEEDINGS{A_TTP,
  author={Chen, Keyan and Liu, Chengyang and Li, Wenyuan and Liu, Zili and Chen, Hao and Zhang, Haotian and Zou, Zhengxia and Shi, Zhenwei},
  booktitle={IGARSS 2024 - 2024 IEEE International Geoscience and Remote Sensing Symposium}, 
  title={Time Travelling Pixels: Bitemporal Features Integration with Foundation Model for Remote Sensing Image Change Detection}, 
  year={2024},
  volume={},
  number={},
  pages={8581-8584},
  doi={10.1109/IGARSS53475.2024.10640593}}

@ARTICLE{A_BiFA,
  author={Zhang, Haotian and Chen, Hao and Zhou, Chenyao and Chen, Keyan and Liu, Chenyang and Zou, Zhengxia and Shi, Zhenwei},
  journal={IEEE Transactions on Geoscience and Remote Sensing}, 
  title={{BiFA}: Remote Sensing Image Change Detection With Bitemporal Feature Alignment}, 
  year={2024},
  volume={62},
  number={},
  pages={1-17},
  doi={10.1109/TGRS.2024.3376673}}

@ARTICLE{A_SAAN,
  author={Guo, Haonan and Su, Xin and Wu, Chen and Du, Bo and Zhang, Liangpei},
  journal={IEEE Transactions on Image Processing}, 
  title={{SAAN}: Similarity-Aware Attention Flow Network for Change Detection With VHR Remote Sensing Images}, 
  year={2024},
  volume={33},
  number={},
  pages={2599-2613},
  doi={10.1109/TIP.2024.3349868}}

@article{A_UNet,
  title={{T-UNet}: triplet UNet for change detection in high-resolution remote sensing images},
  author={Zhong, Huan and Wu, Chen},
  journal={Geo-spatial Information Science},
  volume={28},
  number={2},
  pages={437--454},
  year={2025},
  publisher={Taylor \& Francis}
}

@ARTICLE{A_HF-MCD,
  author={Cai, Luyang and Sun, He and Sun, Xu and Yan, Huanqian and Gao, Lianru},
  journal={IEEE Transactions on Geoscience and Remote Sensing}, 
  title={{HF-MCD}: A Heterogeneous Fusion Framework for Multimodal Change Detection}, 
  year={2025},
  volume={63},
  number={},
  pages={1-15},
  doi={10.1109/TGRS.2025.3606546}}

@ARTICLE{A_MSCD-Net,
  author={Wang, Jian and Xie, Hong and Yan, Li and Zhou, Tingyuan and Wang, Yanheng and Zhang, Jing and Bruzzone, Lorenzo and Atkinson, Peter M.},
  journal={IEEE Transactions on Geoscience and Remote Sensing}, 
  title={{MSCD-Net}: From Unimodal to Multimodal Semantic Change Detection}, 
  year={2025},
  volume={63},
  number={},
  pages={1-17},
  doi={10.1109/TGRS.2025.3591814}}

@ARTICLE{11313649,
  author={Chen, Keyan and Liu, Chenyang and Chen, Bowen and Zhang, Jiafan and Zou, Zhengxia and Shi, Zhenwei},
  journal={IEEE Transactions on Geoscience and Remote Sensing}, 
  title={{RSRefSeg 2}: Decoupling Referring Remote Sensing Image Segmentation With Foundation Models}, 
  year={2026},
  volume={64},
  number={},
  pages={1-20},
  doi={10.1109/TGRS.2025.3647535}}

@article{Li2025AnnotationFreeOS,
  title={Annotation-Free Open-Vocabulary Segmentation for Remote-Sensing Images},
  author={Kaiyu Li and Xiangyong Cao and Ruixun Liu and Shihong Wang and Zixuan Jiang and Zhi Wang and Deyu Meng},
  journal={ArXiv},
  year={2025},
  volume={abs/2508.18067},
  url={https://api.semanticscholar.org/CorpusID:280711754}
}

@misc{li2024semicdvlvisuallanguagemodelguidance,
      title={{SemiCD-VL}: Visual-Language Model Guidance Makes Better Semi-supervised Change Detector}, 
      author={Kaiyu Li and Xiangyong Cao and Yupeng Deng and Jiayi Song and Junmin Liu and Deyu Meng and Zhi Wang},
      year={2024},
      eprint={2405.04788},
      archivePrefix={arXiv},
      primaryClass={cs.CV},
      url={https://arxiv.org/abs/2405.04788}, 
}

@article{LIU2025111355,
title = {Hierarchical Feature Alignment-based Progressive Addition Network for Multimodal Change Detection},
journal = {Pattern Recognition},
volume = {162},
pages = {111355},
year = {2025},
issn = {0031-3203},
doi = {https://doi.org/10.1016/j.patcog.2025.111355},
url = {https://www.sciencedirect.com/science/article/pii/S0031320325000159},
author = {Tongfei Liu and Yan Pu and Tao Lei and Jianjian Xu and Maoguo Gong and Lifeng He and Asoke K. Nandi}
}

@ARTICLE{10891329,
  author={Liu, Tongfei and Zhang, Mingyang and Gong, Maoguo and Zhang, Qingfu and Jiang, Fenlong and Zheng, Hanhong and Lu, Di},
  journal={IEEE Transactions on Image Processing}, 
  title={Commonality Feature Representation Learning for Unsupervised Multimodal Change Detection}, 
  year={2025},
  volume={34},
  number={},
  pages={1219-1233},
  doi={10.1109/TIP.2025.3539461}}

@article{LU2026431,
title = {{UnravelNet}: A backbone for enhanced multi-scale and low-quality feature extraction in remote sensing object detection},
journal = {ISPRS Journal of Photogrammetry and Remote Sensing},
volume = {231},
pages = {431-442},
year = {2026},
issn = {0924-2716},
doi = {https://doi.org/10.1016/j.isprsjprs.2025.11.002},
url = {https://www.sciencedirect.com/science/article/pii/S0924271625004319},
author = {Wei Lu and Hui-Dong Li and Chao Wang and Si-Bao Chen and Chris H.Q. Ding and Jin Tang and Bin Luo}
}

@ARTICLE{10142024,
  author={Lu, Wei and Chen, Si-Bao and Tang, Jin and Ding, Chris H. Q. and Luo, Bin},
  journal={IEEE Transactions on Geoscience and Remote Sensing}, 
  title={A Robust Feature Downsampling Module for Remote-Sensing Visual Tasks}, 
  year={2023},
  volume={61},
  number={},
  pages={1-12},
  doi={10.1109/TGRS.2023.3282048}}

@article{Chen2025BRIGHTAG,
  title={BRIGHT: A globally distributed multimodal building damage assessment dataset with very-high-resolution for all-weather disaster response},
  author={Hongruixuan Chen and Jian Song and Olivier Dietrich and Clifford Broni-Bediako and Weihao Xuan and Junjue Wang and Xinlei Shao and Yimin Wei and Junshi Xia and Cuiling Lan and Konrad Schindler and Naoto Yokoya},
  journal={ArXiv},
  year={2025},
  volume={abs/2501.06019},
  url={https://api.semanticscholar.org/CorpusID:275458616}
}

@article{HE2023103197,
title = {Cross-modal change detection flood extraction based on convolutional neural network},
journal = {International Journal of Applied Earth Observation and Geoinformation},
volume = {117},
pages = {103197},
year = {2023},
issn = {1569-8432},
doi = {https://doi.org/10.1016/j.jag.2023.103197},
url = {https://www.sciencedirect.com/science/article/pii/S1569843223000195},
author = {Xiaoning He and Shuangcheng Zhang and Bowei Xue and Tong Zhao and Tong Wu}
}


\end{sloppypar}
\end{document}